\documentclass[runningheads]{llncs}

\usepackage{eccv}

\usepackage{eccvabbrv}

\usepackage{graphicx}
\usepackage{booktabs}
\usepackage{zebra-goodies}
\usepackage[accsupp]{axessibility}  

\usepackage{hyperref}

\usepackage{orcidlink}

\usepackage{bm}
\usepackage[table]{xcolor}
\usepackage{multirow}
\usepackage {arydshln}

\usepackage[width=122mm,left=12mm,paperwidth=146mm,height=193mm,top=12mm,paperheight=217mm]{geometry}

\begin{document}

\title{Under One Sun: Multi-Object Generative Perception of Materials and  Illumination} 

\titlerunning{Under One Sun}

\author{Nobuo Yoshii\inst{1}\orcidlink{0009-0005-0688-0306} \and
Xinran Nicole Han\inst{2}\orcidlink{0000-0003-4448-330X} \and
Ryo Kawahara\inst{1}\orcidlink{0000-0002-9819-3634} \and \\
Todd Zickler\inst{2}\orcidlink{0000-0002-3853-1558} \and
Ko Nishino\inst{1}\orcidlink{0000-0002-3534-3447}}

\authorrunning{N.~Yoshii et al.}

\institute{$^1$ Kyoto University \quad $^2$ Harvard University \\
\url{https://vision.ist.i.kyoto-u.ac.jp/research/onesun/}}

\maketitle

\begin{figure}
\vspace{-24pt}
  \centering
  \includegraphics[width=\linewidth]{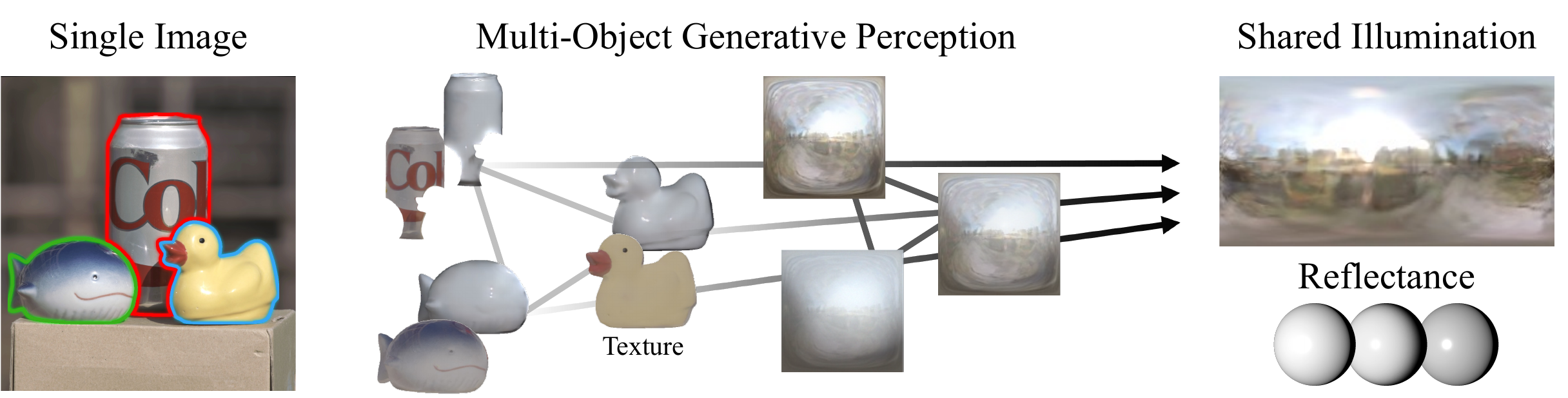}
  \caption{MultiGP is an ambiguity-aware inverse rendering method that samples the reflectance and texture of each object alongside the global scene illumination from a single image. By leveraging the shared illumination across multiple objects, MultiGP employs a novel end-to-end architecture featuring tailored guidance, axial attention, and ControlNet structures to resolve the otherwise ambiguous radiometric disentanglement.}
  \label{fig:teaser}
\end{figure}
\vspace{-24pt}

\begin{abstract}
We introduce Multi-Object Generative Perception (MultiGP), a generative inverse rendering method for stochastic sampling of all radiometric constituents---reflectance, texture, and illumination---underlying object appearance from a single image. Our key idea to solve this inherently ambiguous radiometric disentanglement is to leverage the fact that while their texture and reflectance may differ, objects in the same scene are all lit by the same illumination. MultiGP exploits this consensus to produce samples of reflectance, texture, and illumination from a single image of known shapes based on four key technical contributions: a cascaded end-to-end architecture that combines image-space and angular-space disentanglement; Coordinated Scheduling for diffusion convergence to a single consistent illumination estimate; Axial Attention applied to facilitate ``cross-talk'' between objects of different reflectance; and a Texture Extraction ControlNet to preserve high-frequency texture details while ensuring decoupling from estimated lighting. Experimental results demonstrate that MultiGP effectively leverages the complementary spatial and frequency characteristics of multiple object appearances to recover individual texture and reflectance as well as the common illumination.

\end{abstract}

\section{Introduction}
\label{sec:intro}
The appearance of an object reveals far more than its 3D shape and semantic identity. A glance at a cup, for example, also reveals its material properties (\eg, ceramic, hard, heavy, perhaps with printed text or graphics) and provides informative reflections of the surrounding illumination. Deciphering this rich, entangled information can be essential for interaction. It can allow an embodied agent to plan a grasp based on perceived materials; predict how the object would look from different vantage points; and predict how other objects would look if placed in the same scene.

However, this radiometric disentanglement is overwhelmingly ill-posed. Surrounding light is reflected with an angular response that is unique to the local geometry and reflectance of each surface point, akin to an angular convolution with a shift-dependent kernel~\cite{ramamoorthi2001signal}. This deeply entangles object materials with ambient light, and it is inherently lossy because surface reflectance generally attenuates the light's high angular-frequency details. All of this makes recovering materials and lighting from a single image fundamentally ambiguous~\cite{hartung2002distinguishing}.

``Generative perception'' is a framework for addressing these ambiguities in vision, by using stochastic generative models to sample from the distribution of possible physical explanations. Previous work in this vein has focused on producing samples of material and shape from images~\cite{han2024multistable,zeng2024rgb} or from videos~\cite{DiffusionRenderer,han2025generative}, but without producing samples of illumination. In contrast, DRM \cite{Yenyo_2022_CVPR} produces samples of reflectance \emph{and} illumination from an image of a known shape, but it cannot handle objects with texture. 

In this paper, we show that stochastic sampling of a richer suite of radiometric constituents---reflectance, texture, and illumination---becomes possible when multiple objects are present in the image. 
Our key insight is that while objects in a scene differ in their texture and reflectance, they are all lit by the same illumination.
We introduce Multi-object Generative Perception (MultiGP), an approach that leverages multi-object consensus to produce samples of reflectance, texture and illumination from a single image of known shapes.

MultiGP is built upon four key technical contributions. 
First, we propose a cascaded end-to-end architecture that simultaneously disentangles texture in the image domain and reflectance and illumination in the angular (reflectance map) domain.
Second, we introduce Coordinated Scheduling, which manages the temporal scheduling of the diffusion process to ensure that illumination estimates across different objects converge to one consistent environment map.
Third, we employ Axial Attention to facilitate ``cross-talk'' between objects in the reflectance map domain, allowing objects with different BRDFs to mutually complement each other's missing angular frequencies.
Fourth, we integrate a Texture Extraction ControlNet that preserves high-frequency surface details while ensuring they remain decoupled from the estimated lighting.

Our extensive experiments on synthetic and real-world data demonstrate that MultiGP achieves state-of-the-art accuracy in texture, reflectance, and illumination estimation. 
They also show that MultiGP successfully exploits the complementary information entangled in the object appearance of each object in the spatial and frequency domains. 
Our results suggest that the radiometric information ingrained in object appearance can be reliably exploited by leveraging the physical constraints naturally occurring in multi-object scenes, providing a path toward robust radiometric scene understanding.

\section{Related Works}
There has been considerable recent progress in appearance decomposition using multiple images with changing viewpoint or illumination. Here we focus on single-image methods, where the ambiguity is substantially greater.

\paragraph{\textbf{Deterministic prediction: Inverse rendering and regression.}} Inverse rendering techniques perform gradient descent on a forward rendering operator to recover a set of radiometric constituents that best explain an image, typically using an objective that includes statistical priors or regularization on reflectance and illumination (and sometimes shape when it's unknown)~\cite{lombardi2012reflectance,nishino2009directional,chen2021invertible,yu2023accidental,chen2021dib}. Instead of using iterative optimization, regression-based methods train deep networks to deterministically map an input image to one or more of texture, reflectance or illumination~\cite{wei2020object, maximov2019deep, chen2019learning, rematas2016deep, georgoulis2016delight, georgoulis2017around, meka2018lime}. There are also hybrid methods that use regression to predict some radiometric constituents and inverse rendering to predict others~\cite{wang2025materialist}. Many of these methods only produce smooth, low-frequency lighting due to their regression to the mean and the intrinsic information loss in forward rendering. Also, all of them have the disadvantage of producing a single ``best'' estimate, even when there is fundamental ambiguity, where multiple, diverse explanations are equally valid. In contrast, our method stochastically samples sets of explanatory radiometric constituents, including diverse samples of realistic high-frequency lighting environments that are physically consistent with the observed appearance.

\paragraph{\textbf{Stochastic illumination.}}
Recent generative approaches use diffusion models to hallucinate image-consistent, high-frequency illumination as a panoramic environment map~\cite{shen2025illumidiff, liang2025luxdit} or a virtual mirrored sphere within the scene~\cite{phongthawee2024diffusionlight,liang2024photorealistic}. Although visually plausible, these methods do not explicitly estimate materials, and they fundamentally bypass physically-based modeling of reflection. In contrast, our goal is to radiometrically disentangle object appearance to recover the texture, reflectance, and illumination. DPI~\cite{lyu2023dpi} uses differentiable rendering with a diffusion-based illumination prior.  DRM~\cite{Yenyo_2022_CVPR} considers the special case of textureless objects and reformulates diffusion to generate samples of reflectance and illumination that are consistent with a reflectance map. Our work extends the DRM paradigm by leveraging the common illumination across multiple objects to enable the prediction of texture in addition to reflectance and illumination. 

\paragraph{\textbf{Stochastic texture and reflectance.}} Texture and reflectance estimation has seen significant progress with diffusion-based intrinsic image decomposition \cite{careaga2024colorful,chen2024intrinsicanything,chen2025uni,he2025neural,wang2025mage,sun2025ouroboros,he2025neural,zheng2025dnf}.
Kocsis et al.~\cite{kocsis2024intrinsic} finetune Stable Diffusion to generate samples of texture and reflectance, and they use them to recover illumination by inverse rendering. RGB-X~\cite{zeng2024rgb} and DiffusionRender~\cite{DiffusionRenderer} finetune Stable Diffusion to generate samples of texture, reflectance and shape, but they do not provide illumination.  Recent work by Han et al.~\cite{han2025generative} trains a diffusion model from scratch to generate samples of texture, reflectance, and shape without illumination. In our work, we adapt some components of their U-ViT3D architecture. 

\section{Multi-Object Generative Perception}\label{sec:methods}

\subsection{Problem Formulation}\label{sec:formulation}

An object's appearance can be described by its surface radiance $L_r$
which, ignoring global illumination effects, is determined by the surface
normal $\bm{n}$, the bidirectional reflectance distribution function
(BRDF) $f_r$, and distant incident illumination $L_i$~\cite{kajiya1986rendering}:
\begin{equation}
  L_r(\bm{x},\omega_o) = \int_\omega f_r(\bm{x}, \omega_i, \omega_o)\,
  L_i(\omega_i)\,(\omega_i\cdot\bm{n(x)})\,d\omega_i\,,
  \label{eq:rendering}
\end{equation}
where $\omega_i$ and $\omega_o$ are the incident and outgoing light
directions in the local hemisphere centered at $\bm{n(x)}$. We adopt the isotropic Disney BRDF model~\cite{burley2012physically}:
\begin{align}
    f_r(\bm{x},\omega_i,\omega_o;\bm{\rho_d(x)},\Psi{:=}\{\rho_s,r,\gamma\})
    =(1{-}\gamma)&\frac{\bm{\rho_d(x)}}{\pi}\bigl(f_\text{diff}(\omega_i,\omega_o)
      +f_\text{retro}(\omega_i,\omega_o;r)\bigr) \nonumber \\
    &+f_\text{spec}(\omega_i,\omega_o;\bm{\rho_d(x)},\rho_s,r,\gamma)\,,
    \label{eq:brdf}
\end{align}
where $\bm{\rho_d(x)}$ is the spatially-varying texture and
$\Psi$ is spatially-uniform reflectance consisting of roughness~$r$,
metallicness~$\gamma$, and specular strength~$\rho_s$. Most daily objects exhibit homogeneous specular reflection (\eg, due to a uniform coating over a spatially-varying diffuse reflection). Handling fully spatially-varying specular reflection is left for future work.

Inverting \cref{eq:rendering} for a single object is fundamentally ambiguous; our key insight is to exploit the complementary radiometric information
carried by \emph{multiple} objects to constrain this inversion. Specifically, given an image $I$ containing $M$ objects with known shapes
$\mathbf{S}{=}\{\bm{n(x)}_m\}_{m=1}^M$, we seek the joint posterior over
per-object textures $\mathbf{T}{=}\{\bm{\rho_d(x)}_m\}_{m=1}^M$, per-object
reflectances $\mathbf{R}{=}\{\Psi^{(m)}\}_{m=1}^M$, and shared
illumination~$L$:
\begin{equation}
   p(\mathbf{T}, \mathbf{R}, L \mid I, \mathbf{S})
   \propto p(I \mid \mathbf{S}, \mathbf{T}, \mathbf{R}, L)\,
          p(\mathbf{T}, \mathbf{R}, L \mid \mathbf{S})\,.
\end{equation}

\begin{figure}[t]
  \centering
  \includegraphics[width=\linewidth]{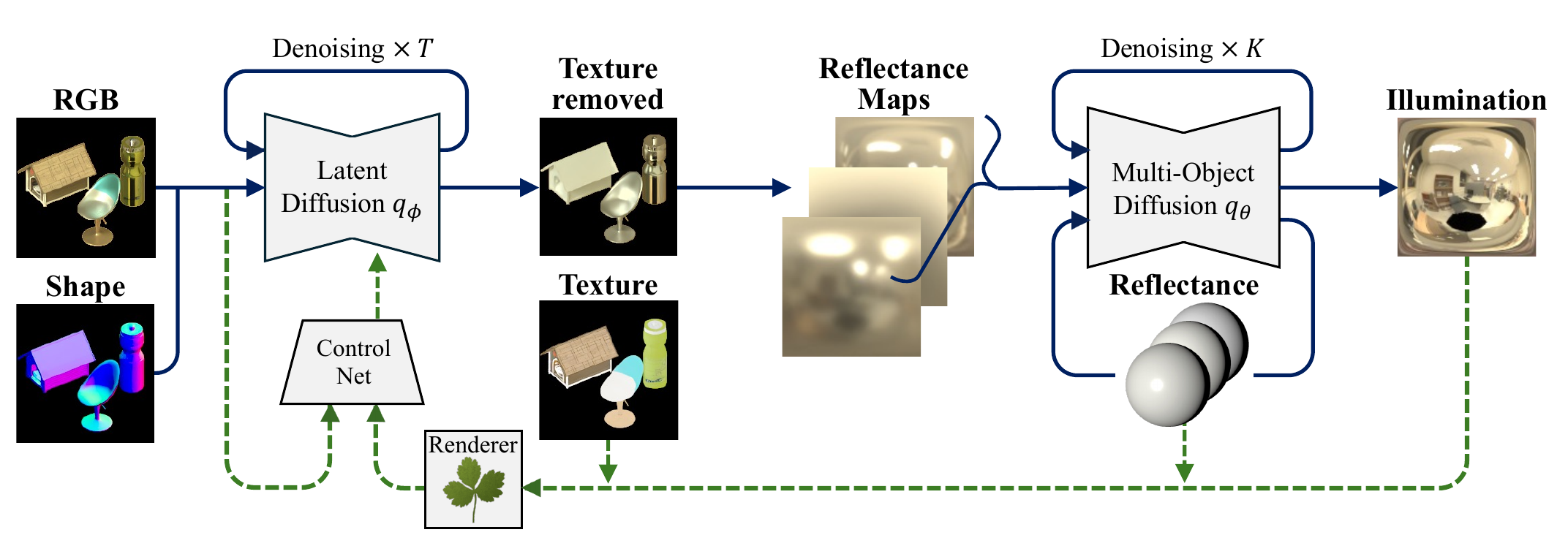}
  \caption{Overview of MultiGP. Given a scene with multiple objects, textures are first estimated via a diffusion model $q_\phi$ that accounts for global light transport. The resulting texture-free appearances are transformed into reflectance maps, from which a multi-object diffusion model $q_\theta$ estimates a shared illumination and respective reflectances. Finally, a ControlNet refines the textures for physical consistency with the estimated lighting and reflectance through a renderer.}
  \label{fig:architecture}
\end{figure}

Computing this posterior exactly is intractable. We approximate it with a
cascaded factorization that separates texture from reflectance and
illumination:
\begin{equation}
    q(\mathbf{T}, \mathbf{R}, L \mid I, \mathbf{S})
    = q_{\phi}(\mathbf{T} \mid I, \mathbf{S})\;
      q_{\theta}(\mathbf{R}, L \mid I, \mathbf{S}, \mathbf{T})\,.
    \label{eq:factorization}
\end{equation}
This factorization directly dictates our architecture
(\cref{fig:architecture}): a texture extraction module $q_\phi$
(\cref{sec:texture_extraction}), followed by a joint
reflectance--illumination module $q_\theta$ that operates on
texture-free, shape-invariant reflectance maps and can therefore focus
entirely on material and lighting estimation
(\cref{sec:multidp-ref}). The estimates are then refined end-to-end by
maximizing the physical likelihood with a ControlNet which examines consistency of the disentanglement and the observation
(\cref{sec:texture_final_refinement}).

\subsection{Texture Extraction Prior ($q_\phi$)}\label{sec:texture_extraction}

The first stage of our factorization isolates the diffuse texture $\mathbf{T}$ from the input image. We employ a latent diffusion model \cite{rombach2022high} conditioned on the concatenated latents of the observed appearance and object shapes, denoted as $\bm{c}$. The intermediate denoising step $t$ follows standard diffusion dynamics:
$$\bm{z_t} = \sqrt{\bar{\alpha}_t}\bm{z_0} + \sqrt{1-\bar{\alpha}_t}\bm{\epsilon},$$
with $\bm{\epsilon} \sim \mathcal{N}(\bm{0}, \bm{I})$. The network $\epsilon_\phi$ predicts the added noise to estimate both the texture $\mathbf{T}$ and the intermediate texture-free appearance $L_r(\bm{x}; \Psi)$. The base loss is defined as 
\begin{equation}
    \mathcal{L}_\phi = \mathbb{E}_{\bm{z_0}, t, \bm{\epsilon}} \left[ \| \bm{\epsilon} - \bm{\epsilon}_\phi(\bm{z_t}, t; \bm{c}) \|_2^2 \right]\,.
\end{equation}

\subsection{Multi-Object Diffusion Reflectance Maps ($q_\theta$)}\label{sec:multidp-ref}

Conditioned on the extracted texture $\mathbf{T}$, we operate on the texture-free appearance of each object to solve $q_{\theta}(\mathbf{R}, L \mid I, \mathbf{S}, \mathbf{T})$. Each texture-free appearance $L_r^{(m)}(\bm{x})$ is mapped to a shape-invariant raw \textit{reflectance map} $\tilde{L}_r^{(m)}(\bm{n})$, a Gaussian sphere of surface radiance indexed by surface normal $\bm{n}$.
They are masked due to the visible normal area of the original objects.
Then, these raw reflectance maps $\tilde{L}_r^{(m)}(\bm{n})$ are converted to complete reflectance maps $L_r^{(m)}(\bm{n})$ via a learned network (a multi-object extension of ObsNet~\cite{Yenyo_2022_CVPR}). This network leverages axial attention analogous to our multi-object axial attention in the diffusion process to mitigate global light transport effects, such as cast shadows and interreflections.

Building on the Diffusion Reflectance Map (DRM) framework~\cite{Yenyo_2022_CVPR}, which recovers illumination from a single object via learned inverse diffusion on the rendering equation, we extend this approach to a multi-object setting. Since each surface reflects different spatial extents and frequency bands of the same illumination, determined by its unique BRDF and geometry, we coordinate these complementary signals to jointly recover the shared illumination $L$ (represented as a mirror reflectance map $L_r^{(0)}(\bm{n})$) and individual reflectances $\mathbf{R}$.

We index each reflectance map by its diffusion step $k$: $L_r^{(m,K)}(\bm{n})$ is the initial observation and $L_r^{(m,0)}(\bm{n}) = \hat{L_i}(\bm{n})$ is the target mirror reflectance map. Following DRM, the forward process is not a Markov noise chain, but a deterministic radiometric formation governed by the rendering equation with additive Gaussian noise to account for the inherent ambiguity. Assuming no cross-talk between object reflections (only the Multi-Object Diffusion $q_\theta$, the Texture Extraction Prior $q_\phi$ and the texture-refinement renderer in \cref{sec:texture_final_refinement} learn from multi-object path-traced data), the joint forward process factorizes over the $M$ objects (see the supplementary material for the full derivation):
\begin{align}
    q(L_r^{(1:M,1:K)} \mid L_r^{(0)},\Psi^{(1:M,K)})
      &= \prod_{m=1}^{M}\prod_{k=1}^{K}
         \mathcal{N}\!\bigl(L_r^{(m,k)} \mid
         L_r(\Psi^{(m,k)}, L_r^{(0)}),\;\sigma^2\bm{I}\bigr)\,,
    \label{eq:our_forward}
\end{align}
where $L_r(\Psi^{(m,k)}, L_r^{(0)})$ denotes the forward rendering operator.

The evidence lower bound (ELBO) decomposes into a sum of per-object terms, yielding the illumination loss:
\begin{align}
    \mathcal{L}_i = \mathbb{E}_{L_i,\Psi,m,k}
      \bigl\lVert \mu_{\theta_i,\theta_r}'
      - \bigl\{L_r(\Psi^{(m,k-1)}, L_r^{(0)}) - L_r^{(m,k)}\bigr\}
      \bigr\rVert_2^2\,,
\end{align}
where $\theta_i$ and $\theta_r$ parameterize the illumination and reflectance estimators, respectively, and $\mu_{\theta_i,\theta_r}'$ is the expectation of the residual between $\hat{L_r}^{(m,k-1)}$ and $L_r^{(m,k)}$. MultiGP minimizes this objective using two key components: \textbf{Coordinated Scheduling}, which guides all object appearances toward a common illumination via reverse diffusion, and \textbf{Axial Attention}, which fuses complementary spatio-spectral information across objects.

\begin{figure}[t]
    \centering
    \begin{minipage}{0.56\linewidth}
        \begin{subfigure}[c]{\textwidth}
        \centering
        \includegraphics[width=\linewidth]{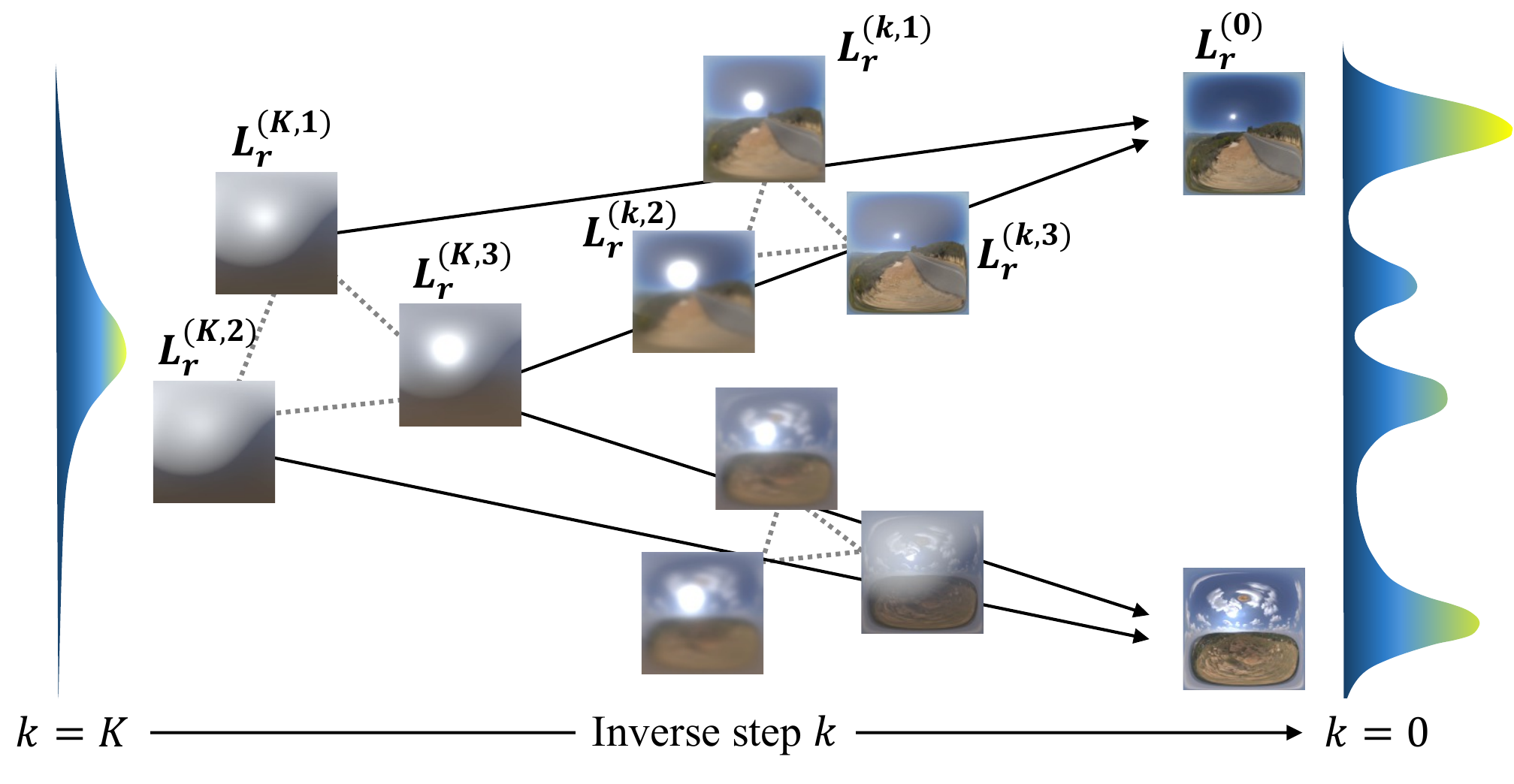}
        \caption{Multi-Object Coordinated Scheduling}
        \label{fig:triangle_step}
        \end{subfigure}
    \end{minipage}
    \hfill
    \begin{minipage}{0.43\linewidth}
        \begin{subfigure}[c]{\textwidth}
        \centering
        \includegraphics[width=\linewidth]{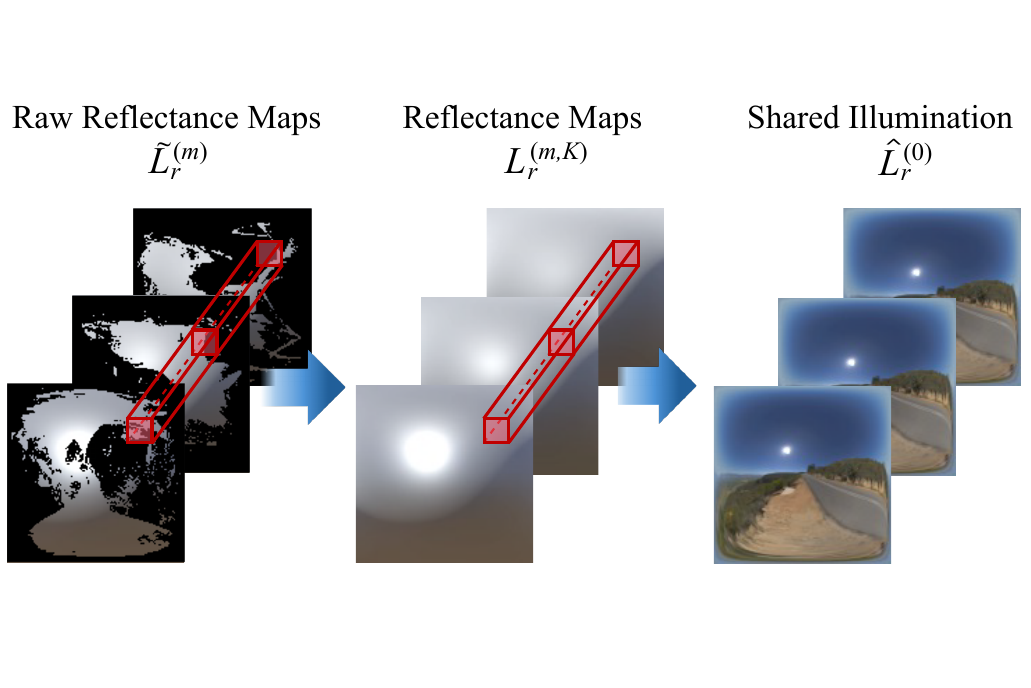}
        \caption{Multi-Object Axial Attention}
        \label{fig:attention}
        \end{subfigure}
    \end{minipage}
    \hfill
    \caption{
 (a) Multi-Object Coordinated Scheduling guides the appearance of diverse objects toward a single shared illumination estimate. At each denoising step, the process is governed by the estimated reflectances of multiple objects alongside a known mirror reflectance. Since the mirror reflectance uniquely corresponds to the illumination, these diverse inputs stochastically converge to a consistent, shared environment map. (b) Multi-Object Axial Attention shares reflectance and spatial information across different reflectance maps. This mechanism enables MultiGP to integrate complementary frequency information which is otherwise attenuated by individual reflectances and aggregate visible lighting directions at every denoising step.
    }
    \label{fig:triangle_attention}
\end{figure}

\paragraph{\textbf{Multi-Object Coordinated Scheduling.}}
Our goal is to convert the $M$ observed reflectance maps $L_r^{(m,K)}$, each representing a different material, back to a single mirror reflectance map $L_r^{(0)}(\bm{n})$ that reveals the environment illumination $\hat{L_i}(\bm{n})$ (\cref{fig:triangle_step}). To achieve this, we schedule the reflectance of each object to evolve linearly from its estimated material $\hat{\Psi}^{(m,K)}$ to the known mirror reflectance $\Psi_0$ over $K$ shared steps:
\begin{align}
    \hat{\Psi}^{(m,k)}=\frac{k}{K}\hat{\Psi}^{(m,K)} + \bigl(1-\frac{k}{K}\bigr)\Psi_0\,.
    \label{eq:our_step}
\end{align}
This ensures that all $M$ objects reach mirror reflectance simultaneously at $k{=}0$ while preserving their relative scales, in contrast to DRM~\cite{Yenyo_2022_CVPR} which requires infinite steps to converge.

The total number of steps $K$ is set proportionally to the average distance of the input materials from mirror reflectance:
\begin{align}
    K = \frac{K_{\text{max}}}{M}\sum^M_{m=1}\Bigl\lVert\frac{\hat{\Psi}^{(m,K)}-\Psi_0}{\sqrt{3}}\Bigr\rVert^2\,,
    \label{eq:our_K}
\end{align}
where $K_{\text{max}}$ is the maximum allowed steps and $\sqrt{3}$ normalizes by the largest possible distance in the 3-D reflectance space $\hat{\Psi}=\{\text{metallic, roughness, specular}\}$. \Cref{eq:our_K} guarantees that the most distant reflectance $\Psi=\{0,1,0\}$ can reach mirror state $\Psi_0=\{1,0,1\}$ in $K_{\text{max}}$ steps.

The model is trained to predict the reflectance at each step with the loss:
\begin{align}
    \mathcal{L}_r = \mathbb{E}_{L_i,f_r,m,k}\bigl[\lVert\hat{\Psi}_{\theta_r}^{(m,k)}-\Psi^{(m,k)}\rVert_2^2+\lVert\hat{\Psi}_{\theta_r}^{(m,K)}-\Psi^{(m,K)}\rVert_2^2\bigr]\,.
\end{align}
The total loss is defined as $\mathcal{L}=\lambda_i\mathcal{L}_i+\lambda_r\mathcal{L}_r$, where $\lambda_i,\lambda_r$ are hyperparameters.

\paragraph{\textbf{Multi-Object Axial Attention.}}
The texture-free reflectance maps $\tilde{L}_r^{(m)}(\bm{n})$ transformed from each object are spatially sparse. Unless the object is a sphere, its geometry does not densely cover the full hemisphere of camera-facing surface normals. More interestingly, different materials act as distinct frequency filters on the environment lights \cite{ramamoorthi2001signal, ramamoorthi2001efficient}: a Lambertian-like surface reflects low-pass filtered illumination in its appearance, while more specular objects retain relatively higher-frequency details through band-pass filtering of the illumination spectrum. No single material behaves as a Dirac delta that preserves all frequencies. The multiple object appearances capture complementary spatio-spectral characteristics of the shared illumination, and we wish to fully leverage this complementarity.

To this end, we propose a Multi-Object Axial Attention mechanism. Inspired by temporal attention in video diffusion~\cite{han2025generative}, this cross-object attention operates across the $M$ reflectance maps at each corresponding spatial location (\ie, the same surface normal direction). As illustrated in Figure (\cref{fig:attention}), the axial attention applied over multiple partially-observed reflectance maps allows an object with missing spectral-frequency details or unobserved normal directions to borrow information from the other objects in the scene, effectively unifying the multi-object observations into a coherent, complete estimate of the global illumination.
We also apply Local Spatial Attention \cite{han2025generative} to avoid information integration with Axial Attention solely along the perfectly aligned normal direction so that MultiGP is made robust to geometry errors.

\subsection{Texture Refinement via ControlNet}\label{sec:texture_final_refinement}

While the cascaded factorization produces diverse samples, some may still violate the physics of image formation. We therefore introduce a final refinement step that injects renderer consistency into the sampling process. Given sampled factors $(\mathbf{T}, \mathbf{R}, L)$ and known geometry $\mathbf{S}$, we render the objects $\hat{L_r}$ with Mitsuba 3 \cite{jakob2022mitsuba3} and compute its residual against the observed image $I$. We condition $q_\Phi$ on this residual via a ControlNet\cite{zhang2023adding}-style mechanism, effectively steering diffusion sampling toward solutions that better explain the observation. The ControlNet accepts the residual $\bm{c_f} = \mathcal{E}(I - \hat{L_r})$ as an additional condition, enforcing tighter physical consistency between texture estimates and global light transport:
\begin{equation}
    \mathcal{L}_{\phi_c} = \mathbb{E}_{\bm{z_0}, t, \bm{\epsilon}} \left[ \| \bm{\epsilon} - \bm{\epsilon}_{\phi_c}(\bm{z_t}, t; \bm{c}, \bm{c_f}) \|_2^2 \right]\,.
\end{equation}
This refinement acts as likelihood-guided sampling rather than explicit optimization; it modifies the sampling dynamics using a renderer-based signal to improve physical consistency while preserving generative diversity.

\section{Experimental Results}

We evaluate MultiGP against several state-of-the-art methods \cite{yu2023accidental, lyu2023dpi, Yenyo_2022_CVPR, phongthawee2024diffusionlight, zeng2024rgb, han2025generative, DiffusionRenderer} on synthetic and real datasets, demonstrating its effectiveness in both accuracy and diversity. Crucially, we derive a canonical framework to evaluate the fundamentally ambiguous disentanglement of inverse rendering components, particularly illumination. We believe this represents a significant contribution to inverse rendering and appearance modeling research at large. Through these extensive evaluations, we show that MultiGP successfully integrates complementary spatio-spectral information from diverse object appearances to achieve superior scene recovery.

\subsection{Dataset}
\paragraph{\textbf{Training.}} We train MultiGP on synthetic 3D multi-object scenes with path-traced light transport (not 2D composites) and per-object reflectance-map triplets rendered by Mitsuba3 \cite{jakob2022mitsuba3}. We create 96 textures for 1100 Adobe 3D Assets \cite{adobestock} and 24 shape-dependent textures for each of the 4000 shapes of Xu \cite{xu2018deep} with MV-Adapter\cite{huang2025mv}. In addition we use unique colored textures.
Reflectance properties (\ie, metallic, roughness, and specular) are randomly sampled, with metallicness restricted to 0 or 1 to avoid unrealistic reflectance under Disney BRDF \cite{burley2012physically}, and roughness set above 0.4 for 50\% of objects to prevent trivially easy illumination estimation. Illumination is randomly sampled from the Laval Indoor Dataset \cite{gardner-sigasia-17} and the Poly Haven HDRIs \cite{polyhaven_hdris} totaling 2000 unique environment maps. 

\vspace{-4pt}
\paragraph{\textbf{Synthetic data.}} For accuracy evaluation, we use the test split of Adobe 3D Assets with MV-Adapter textures and held-out illumination, forming 36 object triplets. For diversity evaluation, we construct pseudo raw reflectance maps from nLMVS-Synth \cite{yamashita2023nlmvs} shapes.

\vspace{-4pt}
\paragraph{\textbf{Real-world data.}} As no existing dataset contains multiple objects in the same image, we construct multi-object evaluation sets from Stanford-ORB \cite{kuang2023stanford} (14 objects across 7 scenes) and nLMVS-Real \cite{yamashita2023nlmvs}. We select triplets of objects captured at approximately the same position, viewpoint, and illumination. This yields 23 and 27 object triplets from Stanford-ORB and nLMVS-Real, respectively.

\vspace{-4pt}
\paragraph{\textbf{MultiGP dataset.}} We additionally capture 9 real scenes (3 outdoor, 6 indoor) featuring 3 objects under shared illumination. Each object's shape was scanned with Artec3D to obtain surface normals and aligned to camera coordinates using MegaPose \cite{labbe2022megapose}.

\vspace{4pt}
\subsection{Ambiguity-Aware Inverse Rendering Metric}

\paragraph{\textbf{Standard Metrics.}} Following \cite{Yenyo_2022_CVPR, phongthawee2024diffusionlight}, we evaluate recovered illumination at 128×256 resolution using scale-invariant logRMSE, PSNR, SSIM, and LPIPS. LogRMSE is computed on HDR images, while other metrics use tone-mapped LDR images. Reflectance is evaluated via scale-invariant logRMSE in the non-parametric MERL BRDF representation \cite{chen2021invertible}. We compare against DPI \cite{lyu2023dpi} and DRM \cite{Yenyo_2022_CVPR}, as methods generating spatially varying BRDFs \cite{zeng2024rgb, han2025generative, DiffusionRenderer} are incompatible with our uniform reflectance setting. Following \cite{kuang2023stanford}, texture is evaluated at 512×512 using scale-invariant RMSE, PSNR, SSIM, and LPIPS.

\paragraph{\textbf{Ambiguity-Aware Metric.}} Standard metrics often falter in inverse rendering because multiple illuminations can produce identical appearances, especially under low-frequency reflectance. To assess how well the estimated distribution captures the ground truth, we utilize Spherical Harmonics (SH). We compute 1089 SH coefficients (up to degree 32) for each illumination sample. By applying PCA (retaining 99\% variance) to the coefficients, we model the sample distribution and report the multivariate normal log-likelihood and Mahalanobis distance of the ground truth within this distribution.

\vspace{-8pt}
\paragraph{\textbf{Evaluation Protocol.}} For all stochastic methods (except ALP \cite{yu2023accidental}), we sample 10 predictions and report the mean of the top 3 \cite{han2025generative}. To evaluate single-object baselines on multi-object scenes, we report results for the object with the sharpest estimated reflectance. DiffusionLight \cite{phongthawee2024diffusionlight} utilizes our reflectance estimate for this selection.

\begin{table}[t]
\begin{center}
\footnotesize
\caption{Quantitative evaluation and ablation study of illumination and reflectance estimates on synthetic test data. For illumination estimation, the sharpest reflectance among multiple objects is shown for both DRM and MultiGP (single). We report the mean of the top 3 out of 10 stochastic predictions. Our full model achieves the highest accuracy and the two novel guidance and attention play crucial roles for this. }
\label{tab:results_synthetic_env}
\resizebox{0.85\textwidth}{!}{
\begin{tabular}{lccccc}
\toprule
& \multicolumn{4}{c}{Illumination} & Reflectance \\
\cmidrule(lr){2-5} \cmidrule(lr){6-6}
& logRMSE$\downarrow$ & PSNR$\uparrow$ & SSIM$\uparrow$ & LPIPS$\downarrow$ & logRMSE$\downarrow$ \\
\midrule
DPI\cite{lyu2023dpi} & 1.64 & 11.96 & 0.37 & \underline{0.56} & 2.22 \\
DRM\cite{Yenyo_2022_CVPR} & 1.48 & 12.62 & 0.34 & 0.61 & 1.85 \\
DiffusionLight\cite{phongthawee2024diffusionlight} & 1.60 & 11.37 & 0.38 & 0.63 & - \\
MultiGP (single) & 1.44 & 12.67 & 0.38 & 0.58 & 2.05 \\
\textbf{MultiGP} & \textbf{1.28} & \textbf{13.54} & \textbf{0.42} & \underline{0.56} & \textbf{1.81}\\
\hdashline
w/o Coordinate Scheduling & \underline{1.29} & \underline{13.50} & \textbf{0.42} & \textbf{0.54} & 1.87\\
w/o Axial Attention & 1.37 & 12.97 & \underline{0.40} & \underline{0.56} & \underline{1.82} \\
w/o Texture Refinement & \underline{1.29} & 13.49 & \textbf{0.42} & \underline{0.56} & 1.87 \\
Global Illumination off & 1.33 & 13.08 & \textbf{0.42} & \underline{0.56} & \textbf{1.81} \\
\bottomrule
\end{tabular}
}
\end{center}
\end{table}

\begin{table}[t]
\begin{center}
\footnotesize
\caption{Quantitative evaluation and ablation study of texture estimation on synthetic test data. MultiGP achieves superior accuracy compared to DPI, which similarly assumes known object geometry.}
\label{tab:results_synthetic_tex}
\resizebox{0.67\textwidth}{!}{
\begin{tabular}{lcccc}
\toprule
&RMSE$\downarrow$ & PSNR$\uparrow$ & SSIM$\uparrow$ & LPIPS$\downarrow$ \\
\midrule
RGB$\leftrightarrow$X \cite{zeng2024rgb} (LDR)  & 0.11 & 27.37 & \underline{0.96} & 0.053 \\
RGB$\leftrightarrow$X \cite{zeng2024rgb} (HDR)  & 0.11 & 27.59 & \underline{0.96} & 0.054 \\
GP Motion \cite{han2025generative} & 0.13 & 25.91 & 0.95 & 0.071 \\
DiffusionRenderer \cite{DiffusionRenderer} & 0.11 & 27.37 & \underline{0.96} & 0.051 \\
\midrule
DPI \cite{lyu2023dpi} & 0.15 & 24.86 & 0.92 & 0.084 \\
\textbf{MultiGP} & \textbf{0.081} & \textbf{29.99} & \textbf{0.97} & \textbf{0.031} \\
\hdashline
w/o Texture Refinement & \underline{0.088} & \underline{29.32} & \textbf{0.97} & 0.034 \\
Global Illumination off & 0.089 & 29.35 & \textbf{0.97} & \underline{0.033} \\
\bottomrule
\end{tabular}
}
\end{center}
\end{table}

\subsection{Results on Synthetic Dataset}

\vspace{4pt}
\paragraph{\textbf{Accuracy.}}\cref{tab:results_synthetic_env} shows quantitative results of the illumination and the reflectance accuracy.
For DRM \cite{Yenyo_2022_CVPR}, which processes only a single object, we report scores for the object with the sharpest reflectance (closest to a mirror state).
MultiGP (single) follows the same protocol. For DiffusionLight \cite{phongthawee2024diffusionlight}, we provide the background consistent with their original assumptions. MultiGP achieves state-of-the-art accuracy across this extensive dataset. \cref{tab:results_synthetic_tex} evaluates texture estimates, where MultiGP outperforms DPI \cite{lyu2023dpi} among methods assuming known geometry.

\paragraph{\textbf{Ablation studies.}}\cref{tab:results_synthetic_env} and \cref{tab:results_synthetic_tex} demonstrate the impact of Coordinate Scheduling, Axial Attention, and Texture Refinement. Our results confirm that Coordinate Scheduling and Axial Attention are critical for estimating high-fidelity illumination and reflectance. Furthermore, Texture Refinement utilizing the estimated individual reflectances and shared illumination significantly improves the accuracy of texture extraction.
Multi-Object Coordinate Scheduling and Axial Attention further attenuate non-shared Global Illumination signals, hence our no-cross-talk assumption in Multi-Object Diffusion. \cref{tab:results_synthetic_env} and \cref{tab:results_synthetic_tex} directly ablates GI in the inputs: GI-on yields 1.28 logRMSE vs. 1.33 for GI-off—MultiGP models real GI.

\paragraph{\textbf{Runtime comparison.}} On one A100 per 1024×1024 multi-object image, MultiGP takes 25.3 seconds, DPI \cite{lyu2023dpi} 1167.7s, DiffusionLight \cite{phongthawee2024diffusionlight} 1280.1s, and DRM \cite{Yenyo_2022_CVPR} 12.1s (single object).
MultiGP is $\sim$50x faster than DPI and DiffusionLight.

\begin{figure}[t]
\vspace{16pt}
  \centering
    \begin{minipage}{0.66\linewidth}
        \centering
        \renewcommand{\thesubfigure}{a-\arabic{subfigure}}
        \begin{subfigure}[c]{0.46\textwidth}
            \centering
            \includegraphics[width=\textwidth]{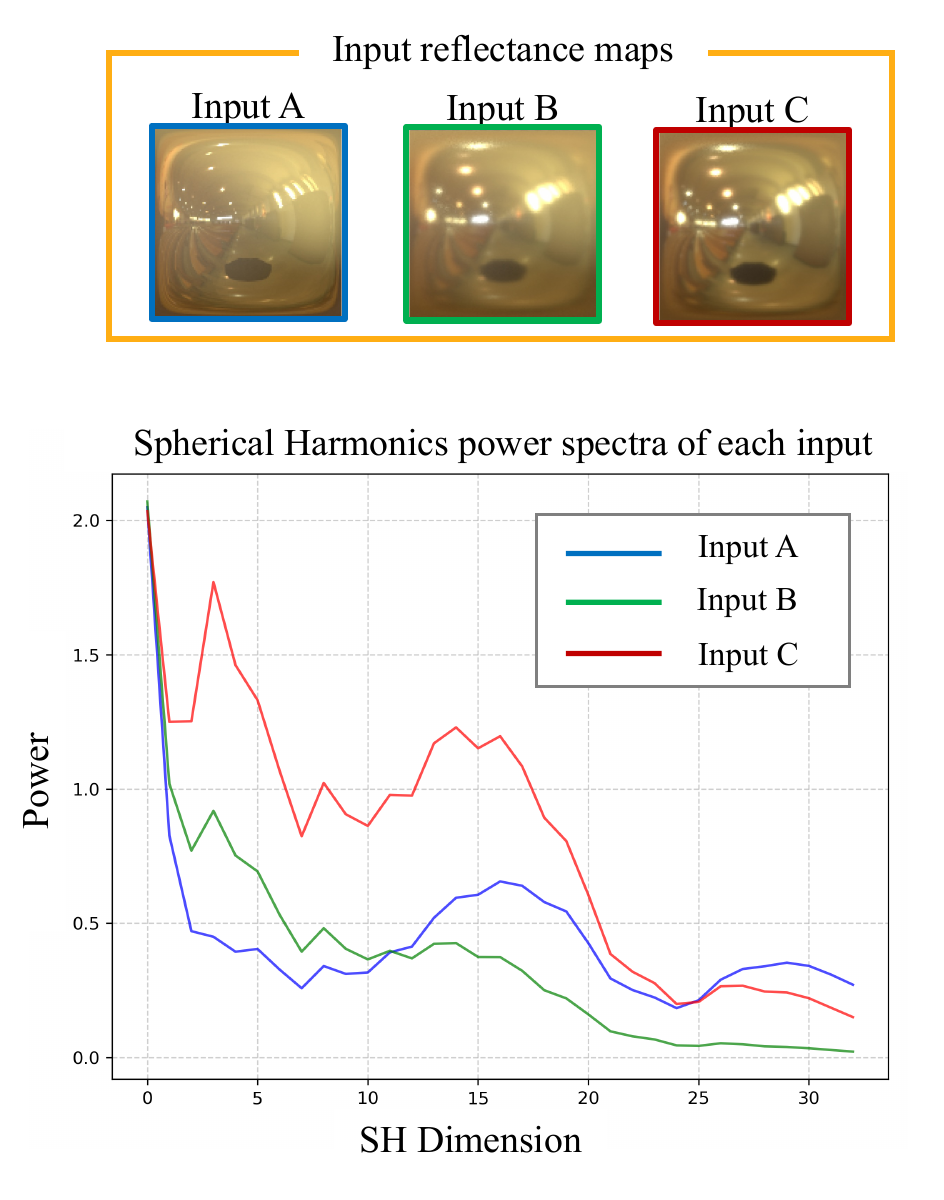}
            \caption{\begin{tabular}[t]{@{}c@{}}Hetero-reflectances\\ (inputs)\end{tabular}}
            \label{fig:hetero_ref_left}
        \end{subfigure}
        \hfill
        \begin{subfigure}[c]{0.52\textwidth}
            \centering
            \includegraphics[width=\textwidth]{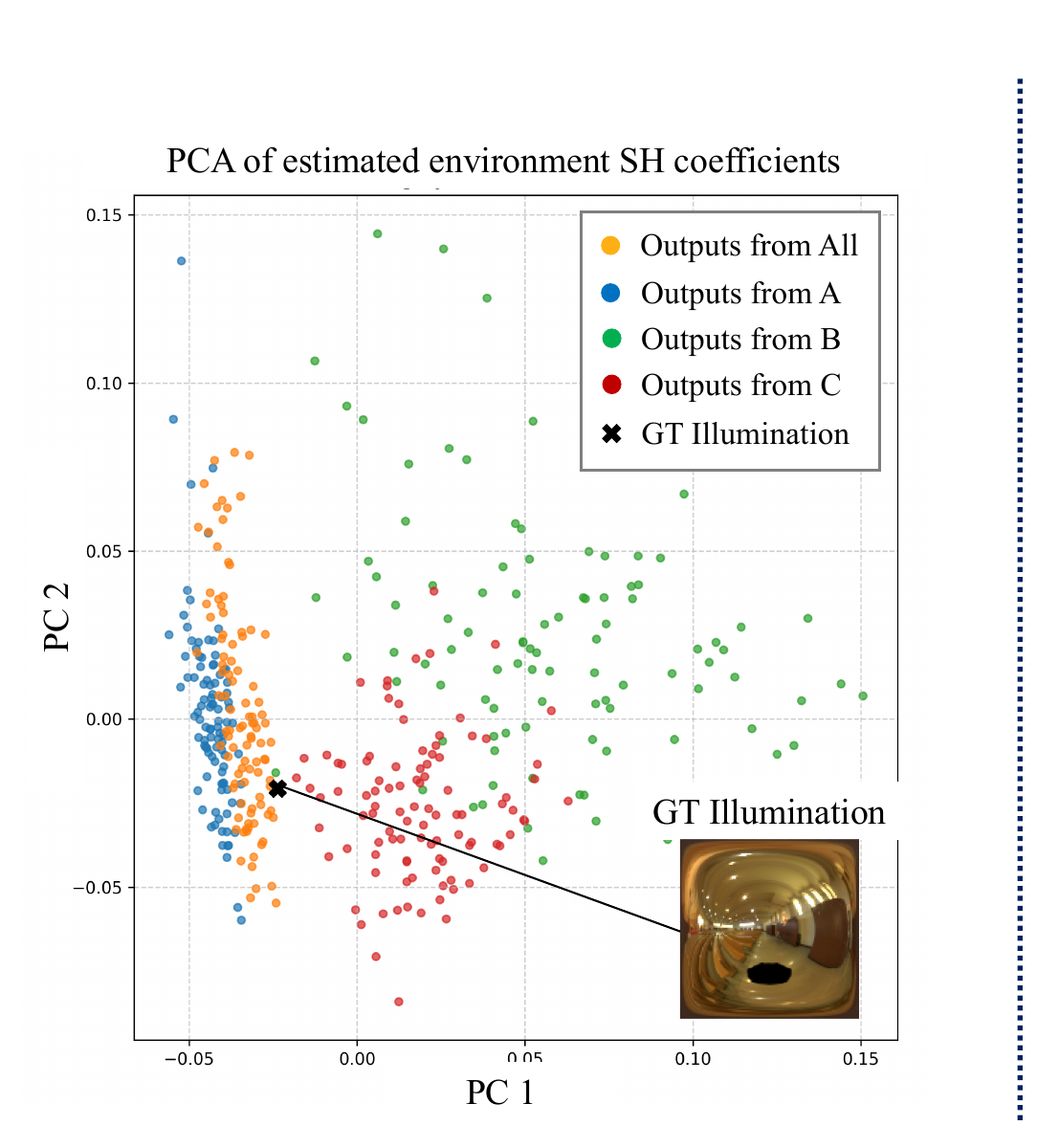}
            \caption{\begin{tabular}[t]{@{}c@{}}Hetero-reflectances\\ (outputs)\end{tabular}}
            \label{fig:hetero_ref_right}
        \end{subfigure}
    \end{minipage}
    \hfill
    \begin{minipage}{0.32\linewidth}
        \centering
        \renewcommand{\thesubfigure}{b}
        \begin{subfigure}[c]{\textwidth}
            \centering
            \includegraphics[width=\textwidth]{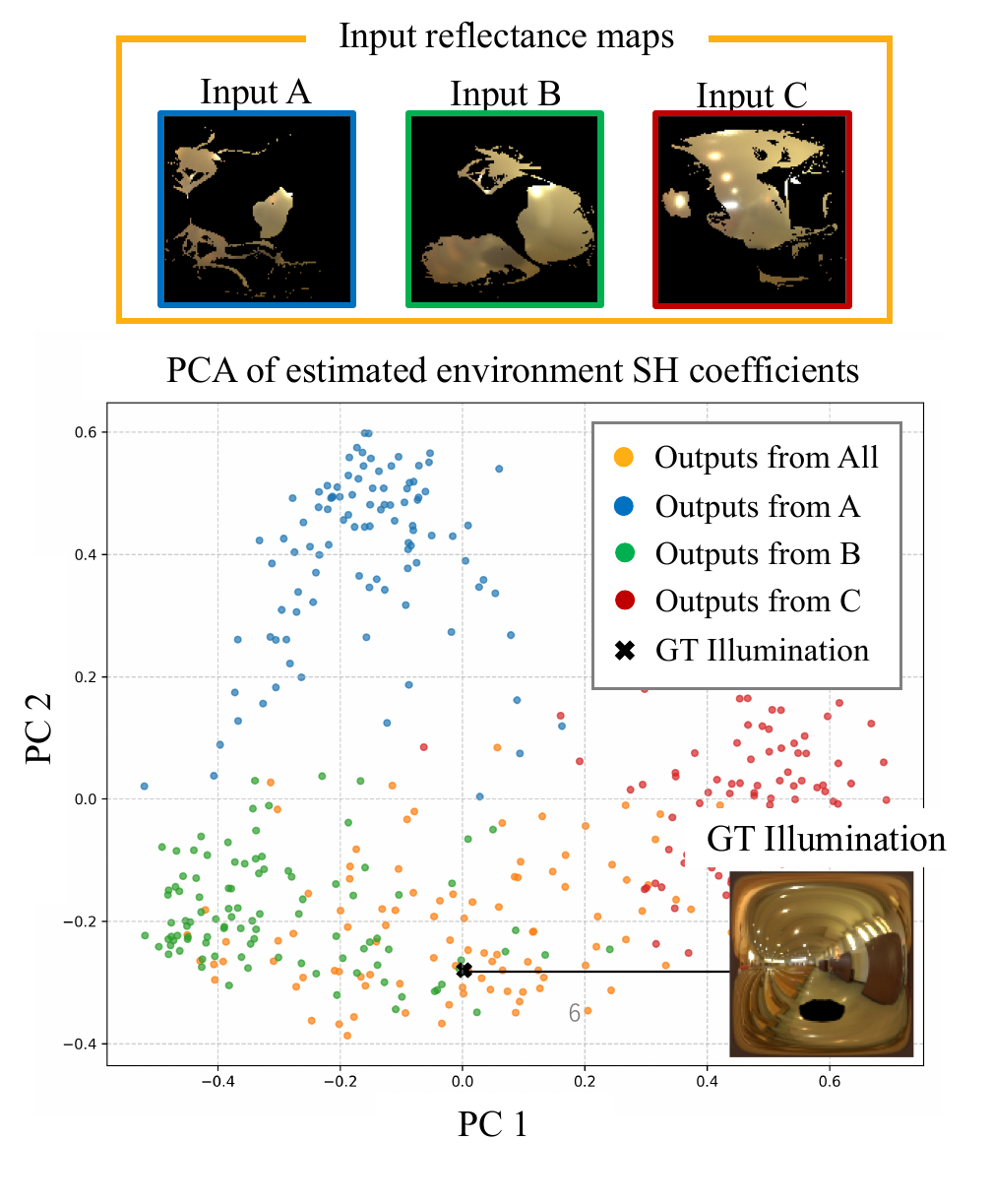}
            \caption{Hetero-masks}
            \label{fig:hetero_mask}
        \end{subfigure}
    \end{minipage}
    \hfill
  \caption{Distribution of illumination samples from MultiGP and MultiGP (single).
(a) Heterogeneous reflectances: MultiGP effectively integrates objects with different reflectances. (a-1) shows the complementary Spherical Harmonic (SH) frequency spectra of three input reflectance maps. (a-2) provides a 2D PCA visualization of 100 samples; orange dots represent the joint MultiGP distribution, while other colors represent individual MultiGP (single) estimates. The joint distribution (orange) captures the ``ground truth'' with the highest density.
(b) Heterogneous masks: Using the same reflectance with different masks demonstrates that MultiGP effectively integrates varied object shapes. Here too, the MultiGP distribution (orange) densely encompasses the ``ground truth.''}
  \label{fig:distribution}
\end{figure}

\vspace{-4pt}
\paragraph{\textbf{Ambiguity-aware accuracy analysis.}}
\cref{fig:distribution} illustrates the distribution of MultiGP estimates. We evaluate MultiGP’s frequency and spatial integration capabilities using two scenarios: (a) objects with varying reflectance (heterogeneous reflectances) and (b) objects with identical reflectance but different geometries (heterogeneous masks).

For objects with distinct frequency characteristics (a), we analyze the Spherical Harmonic (SH) coefficients of the input reflectance maps. As shown in \cref{fig:hetero_ref_left}, input appearances exhibit complementary spectral traits; while the shiniest objects often cover a broad spectrum, they do not provide exhaustive information. Thus, multiple objects collectively offer superior visual cues for illumination recovery compared to any single object.

\cref{fig:hetero_ref_right} visualizes the illumination estimates from 100 MultiGP samples. We compare single-object estimation against our multi-object framework. To facilitate visualization, we apply PCA to the SH coefficients and plot the first two principal components. The multi-object distribution covers the ground-truth illumination with significantly higher likelihood than isolated estimates. This confirms the advantage of leveraging complementary material frequencies. Furthermore, it validates our proposed metric: because inverse rendering is inherently ambiguous, a probabilistic distribution reflects true accuracy more faithfully than a simple distance to ground truth.

\cref{fig:hetero_mask} provides a parallel analysis for scenario (b). The results demonstrate that diverse geometries provide complementary spatial coverage of the environment, which MultiGP successfully exploits to produce more accurate and diverse illumination estimates.

We quantitatively evaluate the distribution of estimates for both cases using synthetic test data (\cref{tab:likelihood}). We randomly paired three object reflectances with 50 test illuminations to render input reflectance maps. For case (b), maps were spatially masked according to visible surface normals from randomly sampled object shapes \cite{yamashita2023nlmvs}. For both scenarios, we sampled 1000 illumination outputs per input and performed PCA on the 1089 order-32 SH coefficients (maintaining a 99\% cumulative eigen-ratio).
Inputs were labeled A, B, and C, corresponding to (a) highest to lowest SH frequency or (b) largest to smallest unmasked area. We then computed the multivariate normal log-likelihood and Mahalanobis distance from the ground-truth illumination within the PCA space.

The results in \cref{tab:likelihood} demonstrate that the highest accuracy is achieved when integrating all objects. In scenario (a), the multi-object accuracy surpasses that of the highest-frequency object (A) alone. Similarly, for spatial integration (b), using all objects outperforms the single largest-coverage shape (A). These results confirm that MultiGP successfully leverages complementary frequency characteristics and spatial spans, justifying a multi-object approach for ambiguity-aware inverse rendering.
\vspace{24pt}

\begin{table}[t]
\begin{center}
\footnotesize
\caption{Likelihood and Mahalanobis distance from ground-truth illumination in PCA space (99\% SH variance), averaged over 50 test scenes with 1000 samples each.
(a) Heterogeneous reflectances: Demonstrates MultiGP's ability to effectively integrate objects with diverse material properties.
(b) Heterogeneous masks: Demonstrates MultiGP's ability to integrate different object shapes and visible areas.
Inputs are labeled A, B, and C, corresponding to (a) highest-to-lowest SH frequency or (b) largest-to-smallest unmasked area. In both cases, the MultiGP ``All'' configuration outperforms the best single-object input (``A'') across all metrics.}
\label{tab:likelihood}

    \begin{minipage}{0.48\linewidth}
        \centering
        \small (a) Heterogeneous reflectances
        \vspace{0.3em}
        \begin{tabular}{ccc}
        \toprule
         Input & \begin{tabular}{c}Multivariate Normal\\ Log-Likelihood $\uparrow$\end{tabular} & \begin{tabular}{c}Mahalanobis\\ Distance $\downarrow$\end{tabular} \\
        \midrule
        All & \textbf{171.9} &  \textbf{16.98} \\
        A & \underline{169.2} &  \underline{18.42}  \\
        B & -137.6 &  26.39  \\
        C & -243.6 &  28.91  \\
        \bottomrule
        \end{tabular}
    \end{minipage}
    \hfill
    \begin{minipage}{0.48\linewidth}
        \centering
        \small (b) Heterogeneous masks
        \vspace{0.3em}
        \begin{tabular}{ccc}
        \toprule
        Input & \begin{tabular}{c}Multivariate Normal\\ Log-Likelihood $\uparrow$\end{tabular} & \begin{tabular}{c}Mahalanobis\\ Distance $\downarrow$\end{tabular} \\
        \midrule
        All & \textbf{61.5} & \textbf{24.18}  \\
        A & \underline{-56.8} &  \underline{27.60}  \\
        B & -696.9 &  38.59  \\
        C & -5211.2 &  57.75  \\
        \bottomrule
        \end{tabular}
    \end{minipage}
    \hfill
    
\end{center}
\vspace{8pt}
\end{table}

\subsection{Results on Real-World Datasets}

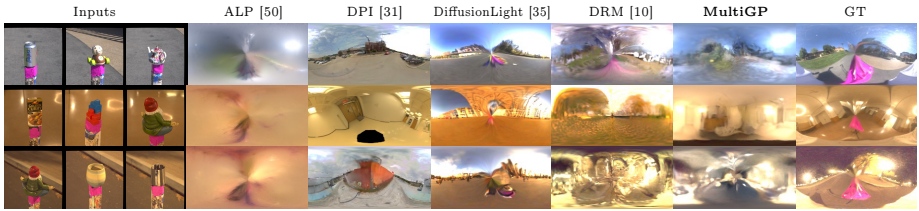
\begin{figure}[t]
    \newcommand{\tinytext}[1]{\tiny\scalebox{0.85}[1.0]{#1}}
    \centering
    \begin{picture}(1\linewidth,8em)
    \put(0,0){\includegraphics[width=\linewidth]{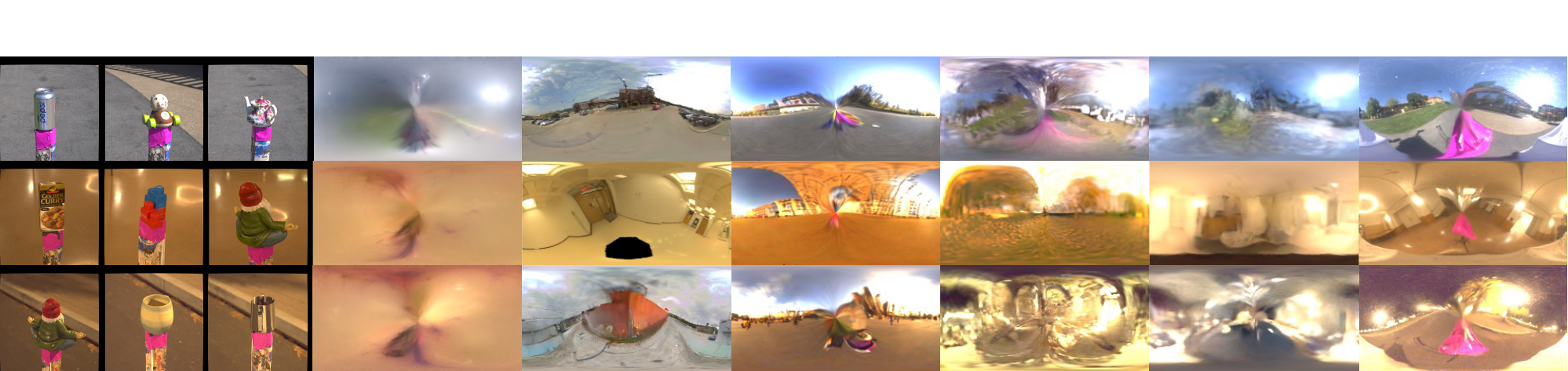}}
    \put(0.075\linewidth,7.8em){\tiny \scalebox{0.85}[1.0]{Inputs}}
    \put(0.24\linewidth,7.8em){\tiny \scalebox{0.85}[1.0]{ALP~\cite{yu2023accidental}}} 
    \put(0.375\linewidth,7.8em){\tiny \scalebox{0.85}[1.0]{DPI~\cite{lyu2023dpi}}} 
    \put(0.47\linewidth,7.8em){\tiny \scalebox{0.80}[1.0]{DiffusionLight\cite{phongthawee2024diffusionlight}} } 
    \put(0.64\linewidth,7.8em){\tiny \scalebox{0.85}[1.0]{DRM~\cite{Yenyo_2022_CVPR}}} 
    \put(0.765\linewidth,7.8em){\tiny \scalebox{0.85}[1.0]{\textbf{MultiGP}}} %
    \put(0.92\linewidth,7.8em){\tiny \scalebox{0.85}[1.0]{GT}} %
    \end{picture}
    \caption{Illumination estimates on the Stanford-ORB dataset. For a fair comparison, we show scaled illumination results closest to the ground truth. For existing methods, we select the result from the object yielding the best logRMSE score. MultiGP faithfully captures the ground truth illumination structure with high-fidelity.}
    \label{fig:figresults_stanfordorb_env}
    \vspace{8pt}
\end{figure}

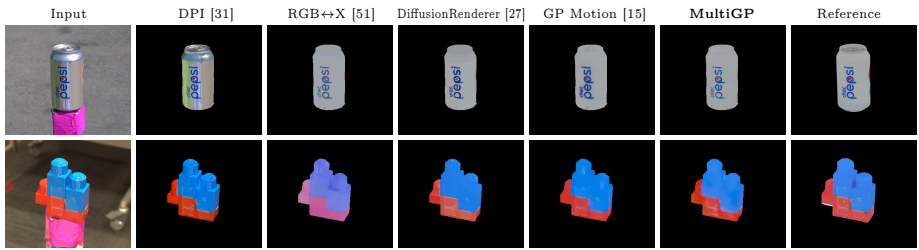
\begin{figure}[t]
    \newcommand{\tinytext}[1]{\tiny\scalebox{0.85}[1.0]{#1}}
    \centering
    \begin{picture}(1\linewidth,9.5em)
    \put(0,0){\includegraphics[width=\linewidth]{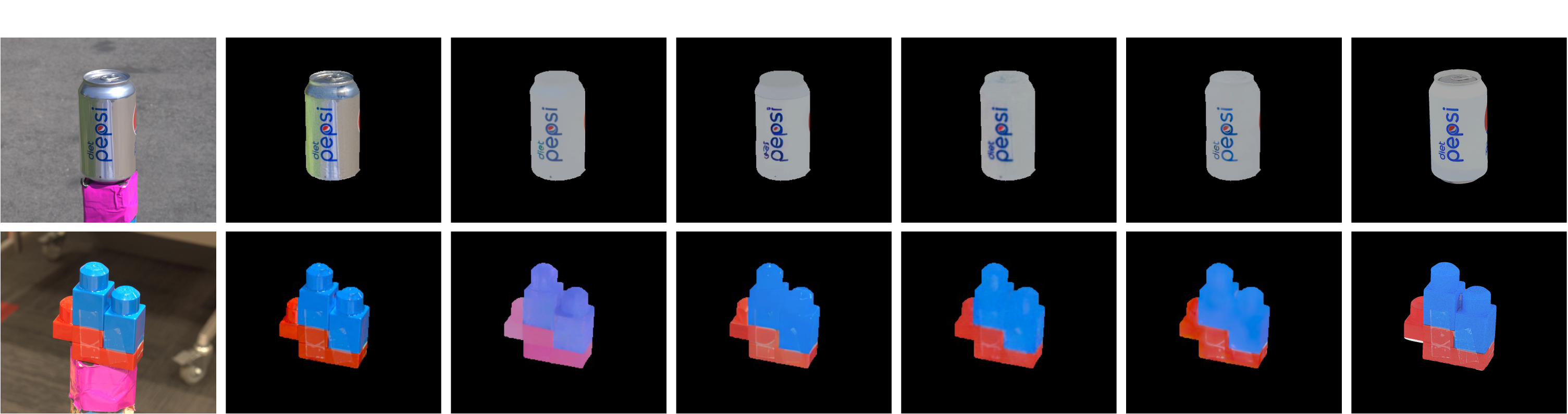}}
    \put(0.05\linewidth,9.3em){\tinytext{Input}}
    \put(0.19\linewidth,9.3em){\tinytext{DPI~\cite{lyu2023dpi}}} 
    \put(0.31\linewidth,9.3em){\tinytext{RGB$\leftrightarrow$X\cite{zeng2024rgb}}} 
    \put(0.43\linewidth,9.3em){\tiny\scalebox{0.73}[1.0]{DiffusionRenderer\cite{DiffusionRenderer}}} 
    \put(0.59\linewidth,9.3em){\tinytext{GP Motion \cite{han2025generative}}} 
    \put(0.75\linewidth,9.3em){\tinytext{\textbf{MultiGP}}}
    \put(0.89\linewidth,9.3em){\tinytext{Reference}}
    \end{picture}
    \caption{Texture estimates on the Stanford-ORB dataset. For fair comparison, scaled albedo that best match the reference within the masked object region are shown. MultiGP estimates highly accurate texture (notice the lack of shading).}
  \label{fig:figresults_stanfordorb_albedo}
\end{figure}

\paragraph{\textbf{Stanford-ORB dataset.}}
\cref{fig:figresults_stanfordorb_env} shows illumination estimates on the Stanford-ORB dataset. Quantitative evaluations are provided in the supplementary material. Ground truth for reflectance is unavailable, as it is nearly impossible to measure for real objects without an integrating (Ulbricht) sphere. While the dataset provides pseudo-ground-truth textures, we found them too noisy for reliable use. Note that ALP requires multi-view images to pre-estimate texture and reflectance, making it non-comparable to our single-image recovery framework. The results for MultiGP (multiple) demonstrate its ability to effectively integrate object appearances for high accuracy. \cref{fig:figresults_stanfordorb_albedo} provides qualitative texture evaluations, showing that MultiGP removes lighting reflections more accurately than other methods, including DPI, which also utilizes surface normals as input.
\vspace{16pt}

\begin{figure}[t]
    \newcommand{\tinytext}[1]{\tiny\scalebox{0.85}[1.0]{#1}}
    \centering
    \begin{picture}(1\linewidth,5.5em)
    \put(0,0){\includegraphics[width=\linewidth]{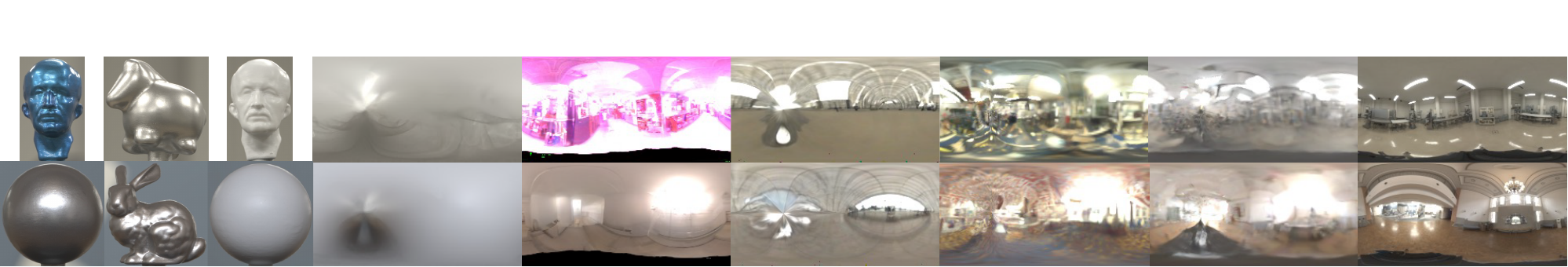}}
    \put(0.075\linewidth,5.3em){\tinytext{Inputs}}
    \put(0.24\linewidth,5.3em){\tinytext{ALP~\cite{yu2023accidental}}} 
    \put(0.375\linewidth,5.3em){\tinytext{DPI~\cite{lyu2023dpi}}} 
    \put(0.47\linewidth,5.3em){\tiny\scalebox{0.80}[1.0]{DiffusionLight\cite{phongthawee2024diffusionlight}} } 
    \put(0.64\linewidth,5.3em){\tinytext{DRM~\cite{Yenyo_2022_CVPR}}} 
    \put(0.765\linewidth,5.3em){\tinytext{\textbf{MultiGP}}} %
    \put(0.92\linewidth,5.3em){\tinytext{GT}} %
    \end{picture}
    \caption{Illumination estimates on the nLMVS-real dataset. For all baselines, the result with best logRMSE is shown. MultiGP results capture the ``ground truth'' light structure most accurately. 
  }
  \vspace{-8pt}
  \label{fig:results_nlmvs}
\end{figure}
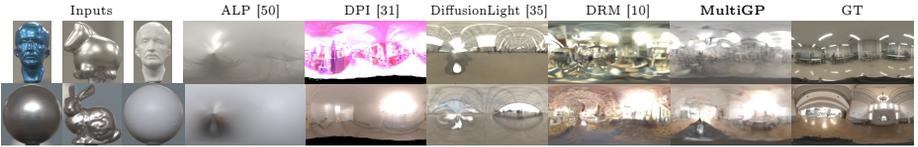

\paragraph{\textbf{nLMVS-real dataset.}}
\cref{fig:results_nlmvs} presents qualitative illumination accuracy on the nLMVS-real dataset. Quantitative results are provided in the supplementary material. All objects in this dataset are textureless, and reflectance ground truth is not available. As before, ALP is shown for reference but is not a competing single-image method, as it requires multi-view images for texture and reflectance pre-estimation. The results demonstrate that MultiGP achieves state-of-the-art accuracy, even for textureless objects, by successfully integrating complementary visual cues of the shared global illumination from surface radiance.

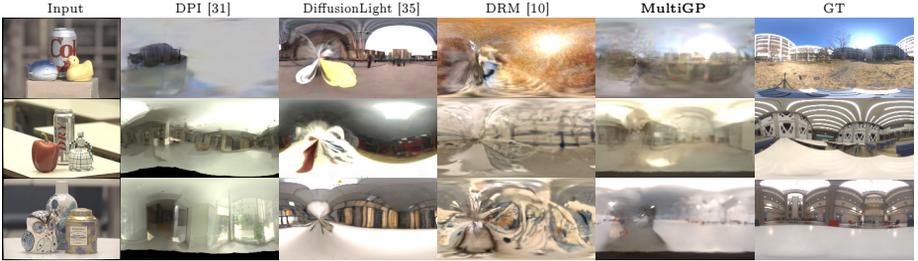
\begin{figure}[t]
    \newcommand{\tinytext}[1]{\tiny\scalebox{0.85}[1.0]{#1}}
    \centering
    \begin{picture}(1\linewidth,10.0em)
    \put(0,0){\includegraphics[width=\linewidth]{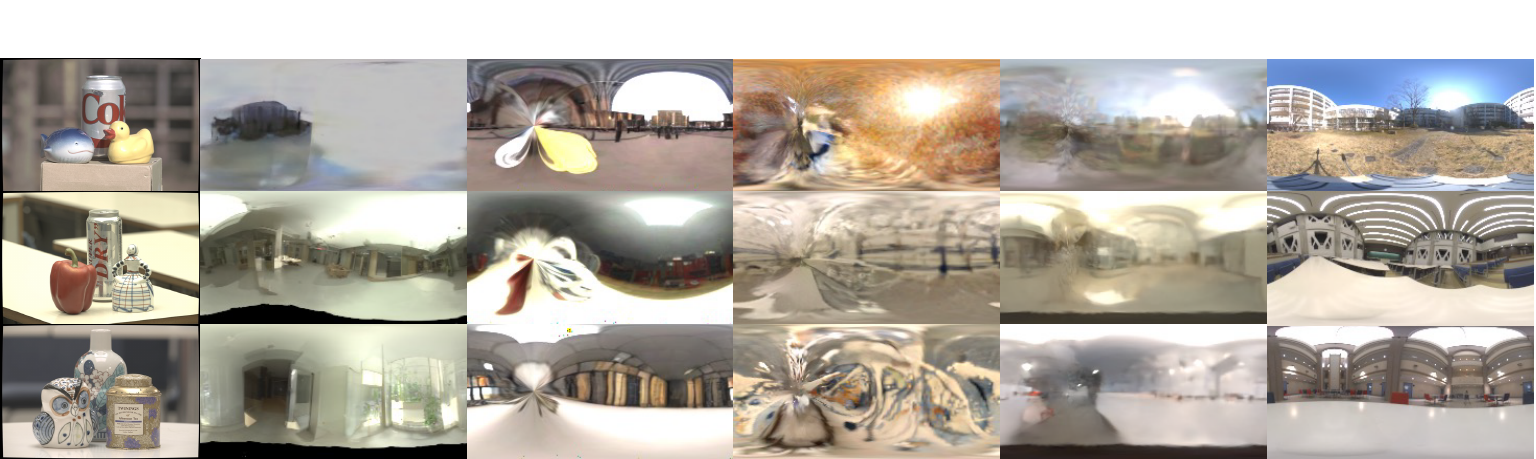}}
    \put(0.05\linewidth,9.99em){\tinytext{Input}}
    \put(0.19\linewidth,9.99em){\tinytext{DPI~\cite{lyu2023dpi}}} 
    \put(0.33\linewidth,9.99em){\tiny\scalebox{0.80}[1.0]{DiffusionLight\cite{phongthawee2024diffusionlight}} } 
    \put(0.53\linewidth,9.99em){\tinytext{DRM~\cite{Yenyo_2022_CVPR}}} 
    \put(0.70\linewidth,9.99em){\tinytext{\textbf{MultiGP}}}
    \put(0.90\linewidth,9.99em){\tinytext{GT}}
    \end{picture}
  \caption{Illumination estimates on newly captured MultiGP dataset. MultiGP accurately recovers the illumination structure of the actual environment despite the complex global light transport.
  }
  \label{fig:results_ourdata_env}
\end{figure}

\paragraph{\textbf{MultiGP dataset.}}
\cref{fig:results_ourdata_env} shows a qualitative evaluation of illumination accuracy on our newly captured multi-object dataset. Quantitative evaluations are provided in the supplementary material. We captured 9 scenes across both indoor and outdoor settings to evaluate multiple objects in a shared physical environment. The results demonstrate that MultiGP generalizes effectively to real-world scenes, successfully handling complex global light transport effects.

\section{Conclusion}

In this paper, we introduced Multi-Object Generative Perception (MultiGP), a framework that resolves the fundamental ambiguities of inverse rendering by leveraging multi-object scene constraints. By treating multiple objects as concurrent probes of the same environment, our method aggregates complementary spatio-spectral information to estimate texture, reflectance, and illumination from a single image. To rigorously evaluate this stochastic approach, we introduced a novel metric for ambiguity-aware inverse rendering, shifting the focus from traditional point estimates to how effectively sampled distributions encompass the physical ground truth. Our results on synthetic and real-world datasets demonstrate that MultiGP significantly outperforms existing methods, providing a robust path toward physics-aware scene understanding.

A primary limitation of our current approach is the requirement for known object geometry, as the radiometric cues are anchored to the 3D surface. Furthermore, our model assumes distant environmental illumination (directional lighting). In real-world indoor scenarios, near-field lighting effects, where light source proximity causes spatially varying illumination, can introduce significant deviations from our assumptions required to model the illumination on the reflectance map. Another limitation of the no-cross-talk assumption in the Multi-Object Diffusion could be relaxed by including explicit interreflection augmentation. Spatially varying reflectance can also be handled by extending the proposed framework with multi-patch reflectance maps: given a material-region segmentation, each region becomes its own ``object'' in our M-input pipeline with its own raw reflectance map, uniform reflectance, and shared illumination. Coordinated Scheduling and Axial Attention operate identically over patches, with more ``objects'' sharing one illumination. Future work will focus on relaxing the geometry constraint through joint shape estimation and extending the generative framework to represent near-field lighting for more complex scenes.

\section*{Acknowledgements}
This work was in part supported by
JSPS KAKENHI
21H04893,
26H02534,
and JST JPMJAP2305.

%
%
\bibliographystyle{splncs04}
\bibliography{main}

\appendix

\section{MultiGP Training Objective}
The illumination loss used as part of the overall loss to train the Multi-object Diffusion Reflectance Maps network is
\begin{align}
    \mathcal{L}_i = \mathbb{E}_{L_i,\Psi,m,k}
      \bigl\lVert \mu_{\theta_i,\theta_r}'
      - \bigl\{L_r(\Psi^{(m,k-1)}, L_r^{(0)}) - L_r^{(m,k)}\bigr\}
      \bigr\rVert_2^2\,.
      \label{eq:our_objective}
\end{align}
We show in this section that this formulation follows from the statistical independence of the diffusion processes for the reflectance maps of the $m=1\ldots M$ observed objects. Because we model the interdependence of multiple object appearances across both frequency and spatial domains via Coordinated Scheduling, we can assume independence in the diffusion objectives. This allows the MultiGP objective to be expressed as a collection of individual DRM~\cite{Yenyo_2022_CVPR} losses.

The DRM forward process assumes additive Gaussian noise within the rendering equation for increasingly attenuated reflectance \cite{Yenyo_2022_CVPR}:
\begin{align}
    q(L_r^{(1:K)} \mid L_r^{(0)},\Psi^{(K)})
      &= \prod_{k=1}^{K}
         q(L_r^{(k)} \mid L_r^{(0)},\Psi^{(K)}) \\
      &= \prod_{k=1}^{K}
         \mathcal{N}\!\bigl(L_r^{(k)} \mid
         L_r(\Psi^{(k)}, L_r^{(0)}),\;\sigma^2\bm{I}\bigr)\,.
    \label{eq:drm_forward}
\end{align}
The MultiGP forward process extends this to $M$ objects as the product of $M$ independent DRM processes:
\begin{align}
    q(L_r^{(1:M,1:K)} \mid L_r^{(0)},\Psi^{(1:M,K)})
      &= \prod_{m=1}^{M}\prod_{k=1}^{K}
         \mathcal{N}\!\bigl(L_r^{(m,k)} \mid
         L_r(\Psi^{(m,k)}, L_r^{(0)}),\;\sigma^2\bm{I}\bigr)\,.
    \label{eq:our_forward}
\end{align}
Correspondingly, the reverse process for $M$ objects is the product of $M$ inverse DRM processes:
\begin{align}
    p_\theta(L_r^{(0:K-1)}|L_r^{(1:M, K)})
    &=\prod^M_{m=1}\prod^K_{k=1}p_{\theta_i,\theta_r}(L_r^{(m, k-1)}|L_r^{(m, k)}) \\
    &=\prod^M_{m=1}\prod^K_{k=1}\mathcal{N}\!\Bigl(L_r^{(m, k-1)}|\mu_{\theta_i,\theta_r}\bigl(L_r^{(m, k)}(\Psi^{(m, k)},k)\bigr),\delta^2\bm{I}\Bigr)\,.
    \label{eq:our_reverse}
\end{align}
Following standard practice \cite{Yenyo_2022_CVPR, rissanen2023generative, ho2020denoising}, we train the model by minimizing negative log-likelihood, $-\log p_\theta$. For a single object, this yields the DRM objective \cite{Yenyo_2022_CVPR}:
\begin{align}
    \mathcal{L}_i = \mathbb{E}_{L_i,\Psi,k}
      \bigl\lVert \mu_{\theta_i,\theta_r}'
      - \bigl\{L_r(\Psi^{(k-1)}, L_r^{(0)}) - L_r^{(k)}\bigr\}
      \bigr\rVert_2^2\,.
    \label{eq:drm_objective}
\end{align}
Since $p_\theta$ in MultiGP is a product across $M$ objects (Eq. \ref{eq:our_reverse}), the total negative log-likelihood simplifies to the sum of $M$ individual DRM objectives. Thus, our objective function \cref{eq:our_objective} directly corresponds to the summation of \cref{eq:drm_objective} over all $m=1\ldots M$ objects.

\section{Training and Inference Details}

\paragraph{Multi-Object Diffusion Reflectance Maps.}
The Multi-Object Diffusion Reflectance Maps network architecture follows U-ViT3D~\cite{han2025generative}, modified to use SwiGLU~\cite{shazeer2020glu} in the transformer blocks.
We employ a lighter version of the U-ViT3D encoder for reflectance by omitting the deepest transformer block.
At each randomly sampled step $k$, the illumination network takes $M$ original and current reflectance maps ($L_r^{(m,K)}$ and $L_r^{(m,k)}$) as input and learns the residual to $L_r^{(m,k-1)}$ via Multi-Object Coordinated Scheduling. Simultaneously, the reflectance network estimates the original reflectance $\Psi^{(m,K)}$, which is converted to the current reflectance $\Psi^{(m,k)}$ using Eq. 8 in the main paper. This estimated reflectance $\Psi^{(m,k)}$ is then integrated into the illumination network as an embedding, similar to a timestep embedding.

We set the reflectance map resolution to $128\times128$. We use the same Mitsuba rendering settings (integrator, sensor, and samples per pixel) as DRM \cite{Yenyo_2022_CVPR}. Textures are set to a uniform white, Training parameters include maximum steps $K_{\text{max}}=150$, with forward and inverse variances set to $\sigma=0.1$ and $\delta=0.125$, respectively. We use loss weights of $\lambda_i=10.0$ and $\lambda_r=0.1$ with a learning rate of $5.0 \times 10^{-5}$.

To stabilize illumination estimation, the first 2,000 epochs are trained using ground-truth reflectance, followed by 2,000 epochs using estimated reflectance. Training takes approximately three days on four A100 GPUs using a large-scale dataset of 1.7 million triplets of pre-rendered reflectance maps. For mutual inpainting of raw reflectance maps, we use an architecture similar to the illumination network but with the channel length and number of heads reduced by one-third. Invisible areas are zero-masked, and we utilize v-prediction rather than the noise masking and $\epsilon$-prediction used in ObsNet \cite{Yenyo_2022_CVPR}. The total pipeline—including texture extraction and refinement—takes 25.3 seconds for a single image containing multiple objects on one A100 GPU.

\begin{figure}[t]
  \centering
  \includegraphics[width=\linewidth]{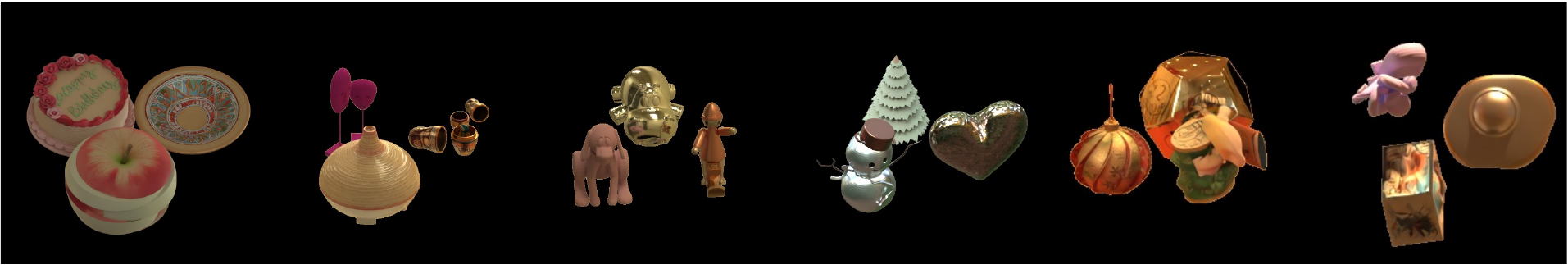}
  \caption{Examples from training data rendered with path-tracing, which includes occlusion, indirect/global-illumination effects, and inter-object shadows.}
  \label{fig:scene_example}
\end{figure}

\paragraph{Texture Extraction and Refinement.}
The architecture is based on the Latent Diffusion Model \cite{rombach2022high}, modified to use v-prediction with a cosine schedule. We set the multi-object path-traced rendering resolution to $256\times256$ and corresponding latent resolution to $64\times64$. Our orthographic rendering pipeline takes either a single object or a triplet (80\% probability) using normalized objects. In the triplet configuration, objects are arranged in an equilateral triangle to prevent interpenetration and subsequently rotated by 15$^\circ$ to 90$^\circ$ vertically and up to 360° horizontally. As shown in \cref{fig:scene_example}, the data includes occlusion,
indirect/global-illumination effects and inter-object shadows. In single-object mode, the asset scale is uniformly sampled between 0.7 and 1.0. To optimize computation, attention is applied only at spatial resolutions of $4\times4$, $8\times8$, and $16\times16$. We fine-tune the pre-trained VQ-VAE (vq-f4) on object images and train the LDM from scratch. Training takes approximately five days on four A100 GPUs.

Following the original ControlNet \cite{zhang2023adding}, our ControlNet is initialized as a copy of the LDM’s learned encoder. Training requires roughly two days on a single H200 GPU. We set the loss weights to $\lambda_{\text{texture}} = \lambda_{\text{texture-free}} = 0.5$ with a learning rate of $5.0 \times 10^{-5}$ and the resolution to $512\times512$ to accurately render fine textures and reflections. During inference, we employ DDIM \cite{song2020denoising} to reduce sampling steps and apply control to the first half of the LDM's layers with a decay factor of 0.825.
Only 50 DDIM steps in one pass is performed for the texture refinement.

\vspace{48pt}
\section{Quantitative Evaluation using Stanford-ORB and nLMVS-real datasets}
\Cref{tab:results_stanfordorb_nlmvs_env} presents the quantitative illumination estimates on the Stanford-ORB and nLMVS-real datasets. Note that ALP \cite{yu2023accidental} requires multi-view images to pre-estimate texture and reflectance, whereas our framework performs recovery from a single image. Despite this difference in input requirements, MultiGP achieves state-of-the-art accuracy across both datasets, delivering performance comparable to---and in several metrics exceeding---ALP. \Cref{tab:results_stanfordorb_tex} presents the “pseudo” quantitative texture estimates on the Stanford-ORB dataset. MultiGP outperforms DPI\cite{lyu2023dpi} among methods assuming known geometry.

\begin{table}[t]
\begin{center}
\footnotesize
\caption{Quantitative evaluation of illumination estimates from real-world data. Results represent the mean of the top 3 out of 10 samples. ALP is included only for reference, as it requires multi-view images for texture and reflectance pre-estimation and is not a direct single-image competitor. MultiGP achieves state-of-the-art accuracy among single-image methods.}
\label{tab:results_stanfordorb_nlmvs_env}
\resizebox{\textwidth}{!}{%
\begin{tabular}{lcccccccc}
\toprule
& \multicolumn{4}{c}{Stanford-ORB dataset} & \multicolumn{4}{c}{nLMVS-real dataset} \\
\cmidrule(lr){2-5} \cmidrule(lr){6-9}
 &logRMSE$\downarrow$\, & PSNR$\uparrow$\, & SSIM$\uparrow$\, & LPIPS$\downarrow$\, &logRMSE$\downarrow$\, & PSNR$\uparrow$\, & SSIM$\uparrow$\, & LPIPS$\downarrow$ \\
\midrule
ALP\cite{yu2023accidental} & \textbf{1.02} & \underline{13.00} & \textbf{0.39} & 0.72 & \underline{0.96} & \underline{13.82} & \underline{0.32} & 0.66\\
\midrule
DPI\cite{lyu2023dpi} & 2.79 & 10.93 & 0.31 & 0.68 & 3.77 & 12.21 & 0.29 & 0.65 \\
DRM\cite{Yenyo_2022_CVPR} & 1.62 & 11.14 & 0.25 & 0.68 & 1.11 & 13.40 & 0.25 & \textbf{0.63} \\
DiffusionLight\cite{phongthawee2024diffusionlight} & 1.49 & 10.99 & 0.33 & \underline{0.63} & 1.39 & 11.75 & 0.28 & 0.65 \\
MultiGP (single) & 1.08 & 12.82 & 0.31 & \underline{0.63} & 1.09 & 12.86 & 0.29 & 0.65 \\
\textbf{MultiGP} & \underline{1.03} & \textbf{13.07} & \underline{0.37} & \textbf{0.62} & \textbf{0.95} & \textbf{14.05} & \textbf{0.33} & \underline{0.64}\\
\bottomrule
\end{tabular}
}
\end{center}
\end{table}

\begin{table}[t]
\begin{center}
\footnotesize
\caption{Quantitative ``pseudo'' evaluation of texture estimates on the Stanford-ORB dataset. Note that the
dataset only provides noisy ``pseudo-ground-truth'' texture. Results represent the mean of the top 3 out of 10 sample. MultiGP achieves superior accuracy compared to DPI, which similarly assumes
known object geometry.}
\label{tab:results_stanfordorb_tex}
\begin{tabular}{lcccc}
\toprule
  & RMSE$\downarrow$ & PSNR$\uparrow$ & SSIM$\uparrow$ & LPIPS$\downarrow$ \\
\midrule
RGB$\leftrightarrow$X\cite{zeng2024rgb}(LDR)  & \underline{0.0265} & \underline{42.23} & \textbf{0.988} & \underline{0.035} \\
RGB$\leftrightarrow$X\cite{zeng2024rgb}(HDR)  & 0.0298 & 41.38 & \underline{0.985} & 0.039 \\
GP Motion \cite{han2025generative} & 0.0284  & 41.66 &  \underline{0.985}  &  0.043 \\
DiffusionRenderer \cite{DiffusionRenderer} & 0.0305 &  41.07  & 0.981 & 0.036 \\
\midrule
DPI\cite{lyu2023dpi} & 0.0441  & 38.28 & 0.974 & 0.041 \\
\textbf{MultiGP} & \textbf{0.0255} & \textbf{42.49} & \underline{0.985} & \textbf{0.032} \\
\bottomrule
\end{tabular}
\end{center}
\end{table}

\vspace{32pt}
\section{MultiGP Dataset Capture}
\begin{figure}[t]
  \vspace{8pt}
  \centering
  \includegraphics[width=\linewidth]{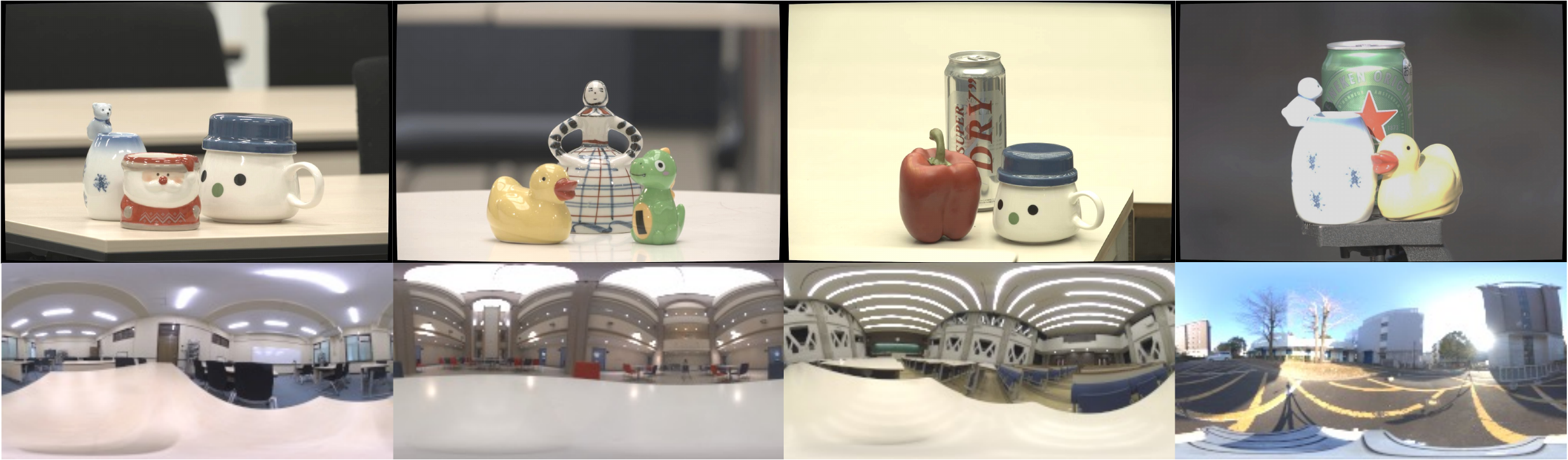}
  \caption{Examples from the newly captured MultiGP dataset. The top row displays input images featuring three objects placed in close proximity, with each at least partially visible. The bottom row shows the corresponding aligned ground-truth illumination maps.}
  \label{fig:capture}
  \vspace{16pt}
\end{figure}

We captured 9 scenes (3 outdoor and 6 indoor), including a courtyard, parking lot, lecture room, and atrium. In each scene, 3 objects were placed in close proximity such that each was at least partially visible from the viewpoint (\cref{fig:capture}); a total of 15 unique objects were used. Ground-truth illumination maps and input images were captured using a RICOH THETA Z1 and a Sony $\alpha7s\mathrm{III}$, respectively. To obtain geometry, each object was scanned with an Artec3D scanner. The resulting surfaces were smoothed to reduce scanning noise, aligned to the captured images using MegaPose \cite{labbe2022megapose} (``megapose-1.0-RGB-multi-hypothesis model", reports 88.6\% GT pose-alignment recall below 5$^\circ$), and converted to surface normal maps via Mitsuba3 \cite{jakob2022mitsuba3} perspective rendering.  Finally, the ground-truth illumination maps were aligned to the camera coordinates using keypoint matching with a chrome ball.

\vspace{16pt}
\section{Quantitative Evaluation using the MultiGP dataset}
\Cref{tab:results_ourdata_env} presents the quantitative results for illumination estimation on the MultiGP dataset. The results demonstrate that MultiGP generalizes effectively to real-world scenes and successfully leverages information from multiple objects to improve performance.
The table also includes experiments evaluating robustness to erroneous geometry using the off-the-shelf surface normal estimation method \cite{garcia2025fine} discussed in \cref{sec:robustness}.

\vspace{16pt}

\begin{table}[t]
\vspace{8pt}
\begin{center}
\footnotesize
\caption{Accuracy of illumination estimates for the newly captured MultiGP dataset. For DRM \cite{Yenyo_2022_CVPR} and MultiGP (single), the object with the sharpest estimated reflectance is selected for evaluation. MultiGP achieves the highest accuracy.} 
\label{tab:results_ourdata_env}
\vspace{-8pt}
\begin{tabular}{lcccc}
\toprule
  & logRMSE$\downarrow$ & PSNR$\uparrow$ & SSIM$\uparrow$ & LPIPS$\downarrow$ \\
\midrule
DPI\cite{lyu2023dpi} & 4.02 & 9.77 & 0.26 & 0.68 \\
DRM\cite{Yenyo_2022_CVPR} & 1.55 & 9.62 & 0.20 & 0.70 \\
DiffusionLight\cite{phongthawee2024diffusionlight} & 1.63 & 10.74 & \underline{0.29} & \textbf{0.64} \\
IID\cite{kocsis2024intrinsic} & 1.36 & 10.00 & \textbf{0.30} & 0.75 \\
IID\cite{kocsis2024intrinsic} (mean-of-10) & 2.29 & 11.00 & 0.28 & 0.67 \\
MultiGP (single) & \underline{1.22} & \underline{11.07} & 0.28 & \underline{0.66}  \\
\textbf{MultiGP} & \textbf{1.14} & \textbf{11.34} & \textbf{0.30} & \underline{0.66}   \\
\hdashline
Marigold-E2E-FT \cite{garcia2025fine} + DRM\cite{Yenyo_2022_CVPR} & 1.67 & 9.58 & 0.19 & 0.69 \\
StableNormal \cite{ye2024stablenormal} + DRM\cite{Yenyo_2022_CVPR} & 1.50 & 9.96 & 0.20 & 0.69 \\
Marigold-E2E-FT \cite{garcia2025fine} + \textbf{MultiGP} & \underline{1.23} & \underline{11.03} & \textbf{0.30} & \underline{0.66}\\
StableNormal \cite{ye2024stablenormal} + \textbf{MultiGP} & \textbf{1.15} & \textbf{11.72} & \underline{0.28} & \textbf{0.66} \\
\bottomrule
\end{tabular}
\end{center}
\end{table}

\vspace{16pt}
\section{Diversity of Illumination Samples}

\begin{figure} 
  \centering
  \includegraphics[width=\linewidth]{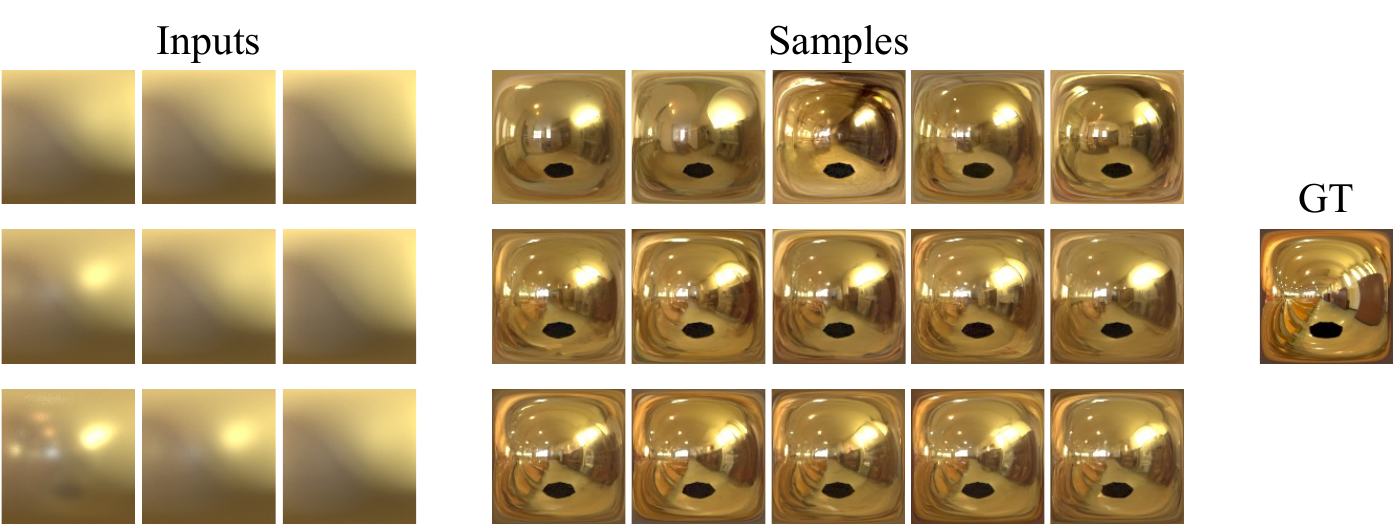}
  \caption{Five illumination samples generated by MultiGP for each input of heterogeneous reflectance. When the input includes shinier objects, the samples are more concentrated near the  ``ground truth.''}
  \label{fig:diverse_heteroref}
\end{figure}

\begin{figure} 
  \centering
  \includegraphics[width=\linewidth]{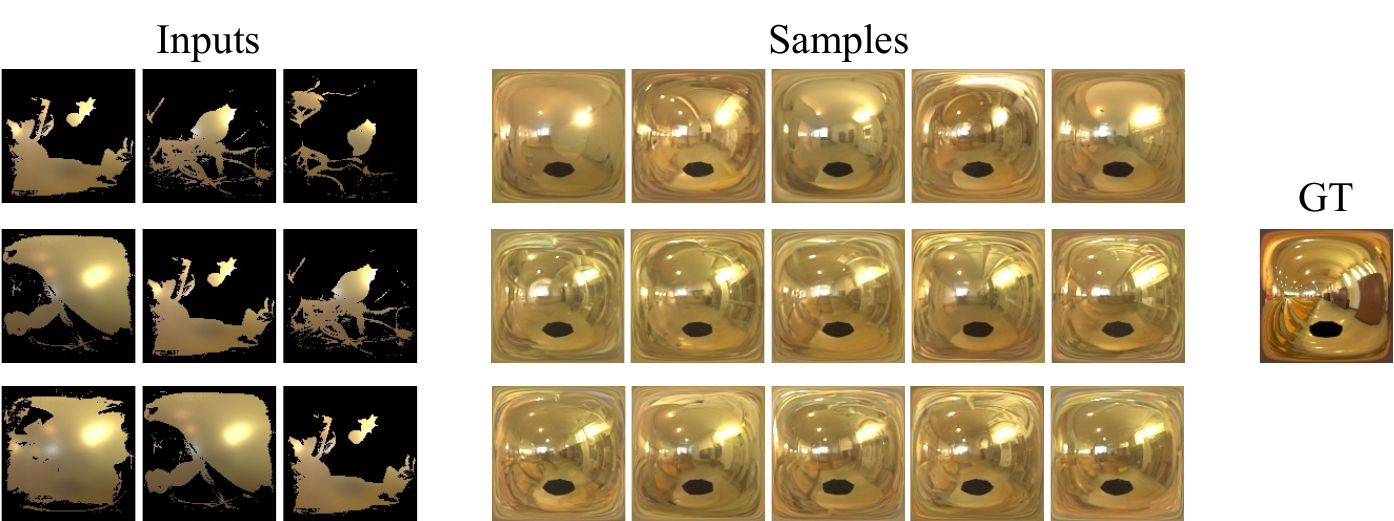}
  \caption{Five illumination samples generated by MultiGP for each input of heterogeneous reflectance. When the input contains larger visible regions, the output samples are more concentrated near the ``ground truth.''}
  \label{fig:diverse_heteromask}
\end{figure}

\Cref{fig:diverse_heteroref,fig:diverse_heteromask} show our model's generated illumination samples for given inputs using different noise seeds. \Cref{fig:diverse_heteroref} demonstrates that illumination samples are more concentrated near the ground truth when the input triplet contains shinier objects. Similarly, \cref{fig:diverse_heteromask} shows that samples become more concentrated when the input contains larger visible regions.

\vspace{32pt}
\section{Fewer Objects}
\begin{figure}[t]
  \centering
    \begin{minipage}{0.66\linewidth}
        \centering
        \renewcommand{\thesubfigure}{a-\arabic{subfigure}}
        \begin{subfigure}[c]{0.46\textwidth}
            \centering
            \includegraphics[width=\textwidth]{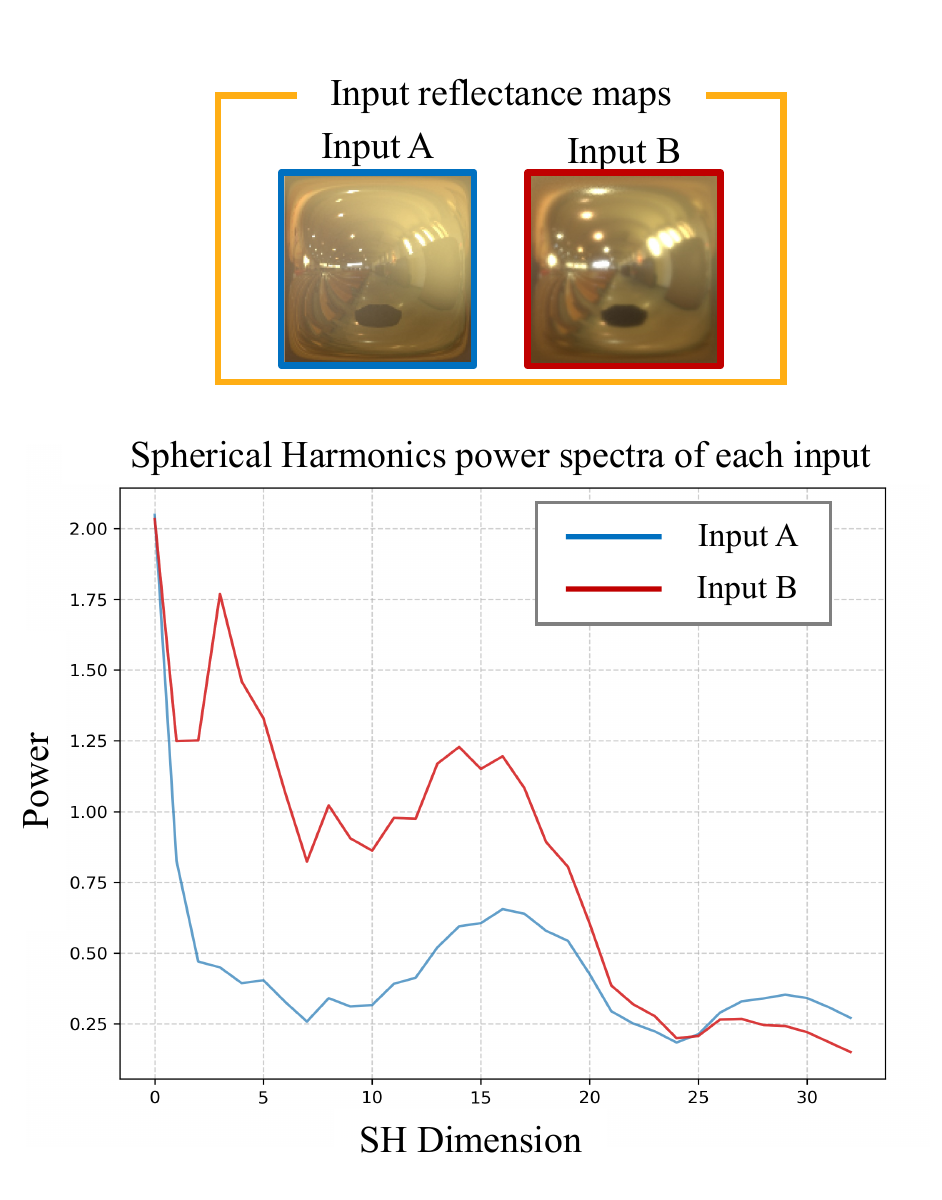}
            \caption{\begin{tabular}[t]{@{}c@{}}Hetero-reflectances\\ (inputs)\end{tabular}}
            \label{fig:hetero_ref_2obj_left}
        \end{subfigure}
        \hfill
        \begin{subfigure}[c]{0.52\textwidth}
            \centering
            \includegraphics[width=\textwidth]{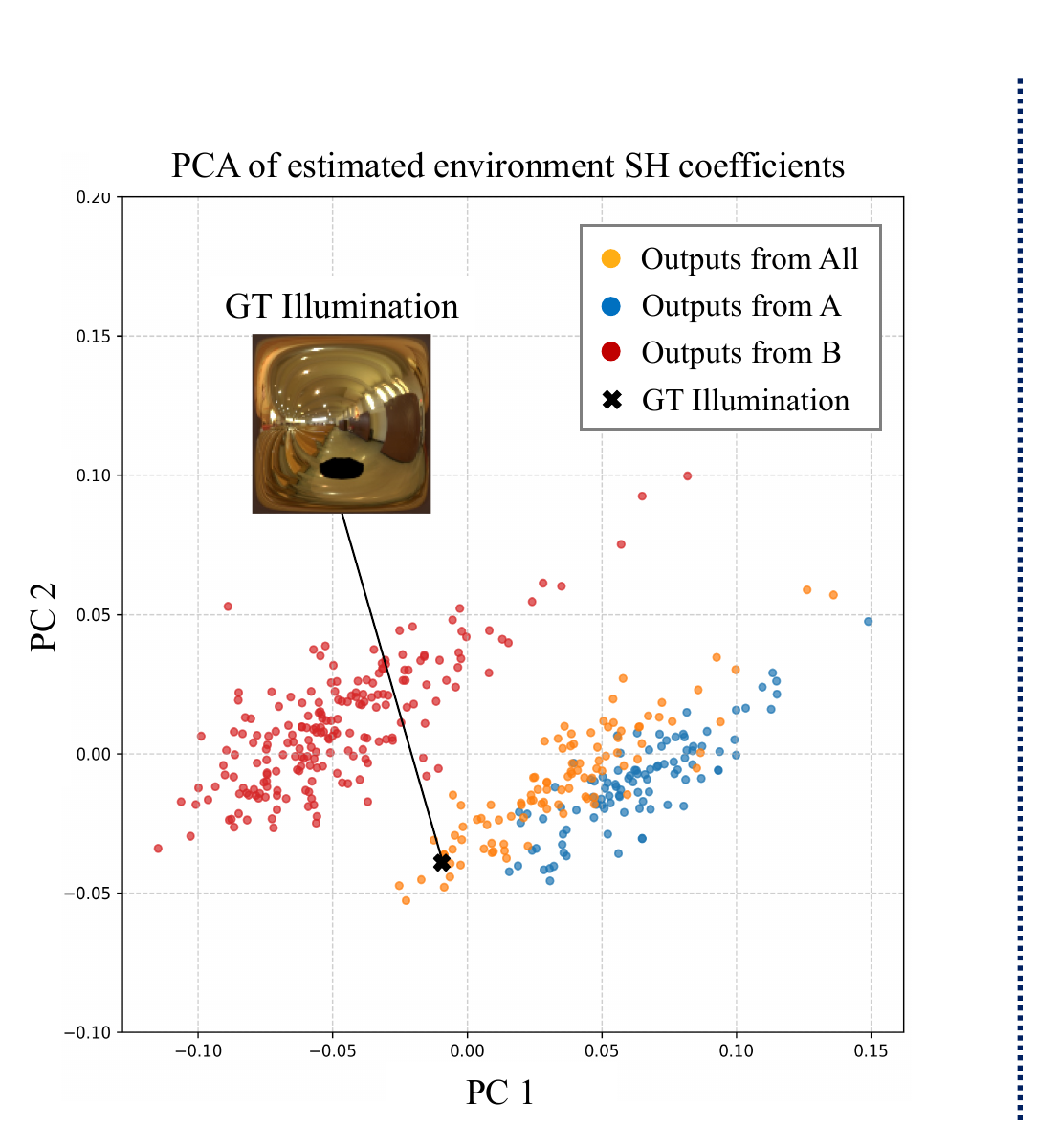}
            \caption{\begin{tabular}[t]{@{}c@{}}Hetero-reflectances\\ (outputs)\end{tabular}}
            \label{fig:hetero_ref_2obj_right}
        \end{subfigure}
    \end{minipage}
    \hfill
    \begin{minipage}{0.32\linewidth}
        \centering
        \renewcommand{\thesubfigure}{b}
        \begin{subfigure}[c]{\textwidth}
            \centering
            \includegraphics[width=\textwidth]{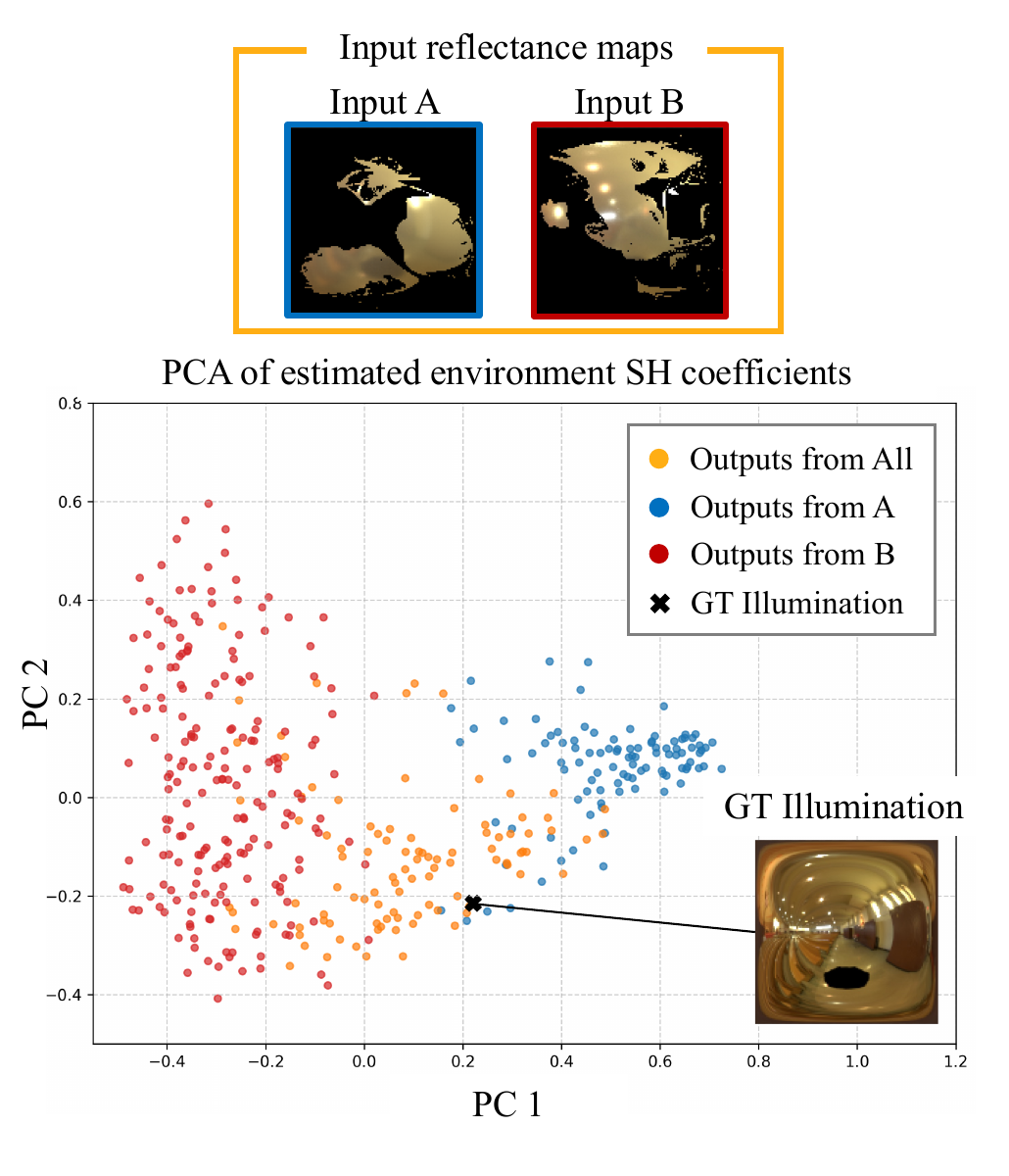}
            \caption{Hetero-masks}
            \label{fig:hetero_mask_2obj}
        \end{subfigure}
    \end{minipage}
    \hfill
  \caption{Distribution of illumination samples from MultiGP and MultiGP (single) when only two input objects are available.
(a) and (b) show heterogeneous reflectances and heterogeneous masks, respectively, analogous to Fig.~4 in the main paper.}
  \label{fig:distribution_nonthree}
\end{figure}

Our model can process fewer than $M$ objects by filling some of the $M$ input channels of our  Multi-object DRM network with duplicates of the reflectance maps that are available.
\Cref{fig:distribution_nonthree} illustrates the distribution of illumination samples from MultiGP and MultiGP (single) when only two objects are available, fewer than the $M=3$ case on which the model was trained.
Specifically, \cref{fig:hetero_ref_2obj_left,fig:hetero_ref_2obj_right} show the input and results for two objects with heterogeneous reflectances, while \cref{fig:hetero_mask_2obj} examines two objects with heterogeneous masks.
In both cases, the output of MultiGP is distributed around ``ground truth'' with higher density than each MultiGP (single) as same as triplet input case in the main paper.

\section{Robustness to Erroneous Geometry}
\label{sec:robustness}
\begin{figure}[t]
  \centering
  \includegraphics[width=\linewidth]{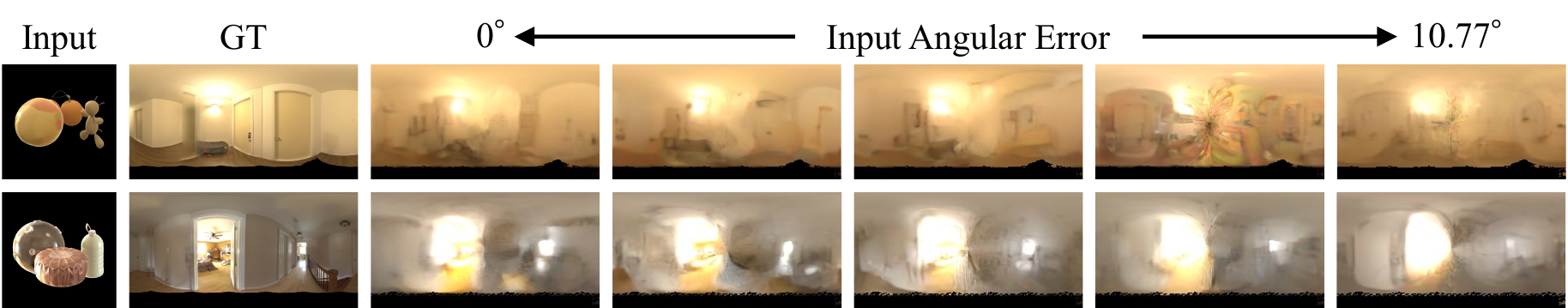}
  \caption{Estimated illuminations become smoother as the noise level of input surface normals increases. However, prominent lighting structures are maintained.}
  \label{fig:angular_robustness}
\end{figure}

\Cref{tab:angular_robustness} and \cref{fig:angular_robustness} present a quantitative evaluation of the effects of noisy surface normal inputs on the quality of illumination estimates. We compare MultiGP with DRM \cite{Yenyo_2022_CVPR} across varying levels of angular noise (AE). 
Specifically, the noisy normals are generated by adding independent gaussian noise $\mathcal{N}(0, \sigma^2)$ to each component of the ground-truth normals, and then scaling them back to unit vectors.
The results indicate that MultiGP maintains higher stability and accuracy under degradation of surface normals, likely benefiting from the complementary information provided by multiple objects, whereas the single-object performance of DRM declines more significantly as input noise increases.
MultiGP further outperforms baselines (Tab.~1 in the main paper) at 18$^\circ$/25$^\circ$ MAE.
\vspace{24pt}

The quantitative results on the real MultiGP dataset (\cref{tab:results_ourdata_env}) include DRM and MultiGP evaluations using the off-the-shelf surface normal estimation method \cite{garcia2025fine, ye2024stablenormal}. While estimation accuracy decreases with erroneous geometry, MultiGP still achieves state-of-the-art performance compared to other methods, including DiffusionLight, which does not require geometry as input.

\begin{table}[t]
\begin{center}
\footnotesize
\caption{Quantitative evaluation of illumination estimates on synthetic test data with noisy surface normal inputs. We report the mean of the top 3 out of 10 stochastic predictions. MultiGP demonstrates robustness to surface normal errors, especially compared to the baseline.}
\label{tab:angular_robustness}
\begin{tabular}{l@{\hspace{8pt}}@{\hspace{4pt}}cccc}
\toprule
\textbf{Method} & \multirow{2}{*}{logRMSE$\downarrow$} & \multirow{2}{*}{PSNR$\uparrow$} & \multirow{2}{*}{SSIM$\uparrow$} & \multirow{2}{*}{LPIPS$\downarrow$} \\
+ Noise Std.($\sigma$) / AE \hspace{4pt} &&&&\\
\midrule
\textbf{MultiGP} & \textbf{1.28} & \textbf{13.53} & 0.42 & \underline{0.56} \\
+ 0.01 / \hspace{4pt}0.72\textdegree & \textbf{1.28} & \underline{13.51} & 0.42 & \textbf{0.55} \\
+ 0.05 / \hspace{4pt}3.59\textdegree & \textbf{1.28} & 13.50 & 0.42 & \underline{0.56} \\
+ 0.10 / \hspace{4pt}7.18\textdegree & \underline{1.32} & 13.21 & 0.42 & 0.57 \\
+ 0.15 / 10.77\textdegree & 1.34 & 13.00 & 0.42 & 0.57 \\
+ 0.25 / 17.95\textdegree & 1.36 & 12.91 & 0.41 & 0.58 \\
+ 0.35 / 25.13\textdegree & 1.38 & 12.63 & 0.41 & 0.59 \\
\midrule
\textbf{DRM} \cite{Yenyo_2022_CVPR} & 1.48 & 12.62 & 0.34 & 0.61 \\
+ 0.01 / \hspace{4pt}0.72\textdegree & 1.54 & 12.59 & 0.35 & 0.61 \\
+ 0.05 / \hspace{4pt}3.59\textdegree & 1.51 & 12.44 & 0.33 & 0.62 \\
+ 0.10 / \hspace{4pt}7.18\textdegree & 1.60 & 12.11 & 0.32 & 0.63 \\
+ 0.15 / 10.77\textdegree & 1.62 & 11.82 & 0.31 & 0.64 \\
+ 0.25 / 17.95\textdegree & 1.66 & 11.38 & 0.31 & 0.65 \\
+ 0.35 / 25.13\textdegree & 1.68 & 11.29 & 0.29 & 0.66 \\
\bottomrule
\end{tabular}
\end{center}
\end{table}

\section{Comparison with Kocsis et al., CVPR 2024 \cite{kocsis2024intrinsic}}
\begin{figure}[t]
  \newcommand{\tinytext}[1]{\tiny\scalebox{0.85}[1.0]{#1}}
  \centering
    \begin{picture}(1\linewidth,6.0em)
    \put(0,0){\includegraphics[width=\linewidth]{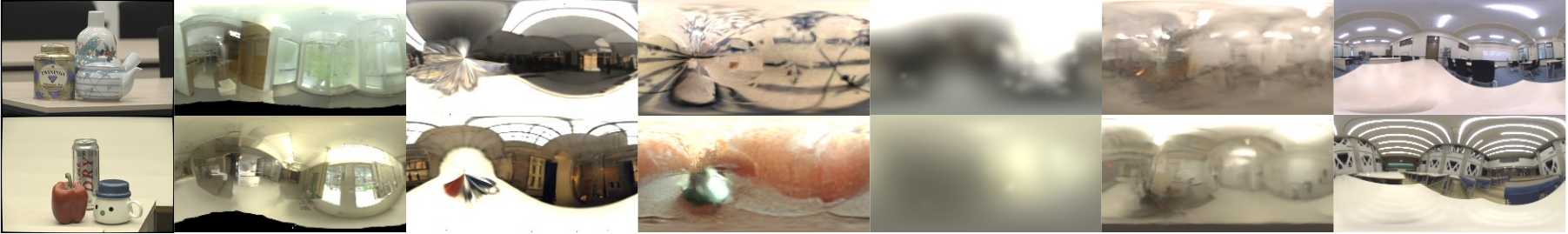}}
    \put(0.035\linewidth,5.8em){\tinytext{Input}}
    \put(0.16\linewidth,5.8em){\tinytext{DPI~\cite{lyu2023dpi}}} 
    \put(0.27\linewidth,5.8em){\tiny\scalebox{0.80}[1.0]{DiffusionLight\cite{phongthawee2024diffusionlight}} } 
    \put(0.455\linewidth,5.8em){\tinytext{DRM~\cite{Yenyo_2022_CVPR}}} 
    \put(0.61\linewidth,5.8em){\tinytext{IID~\cite{kocsis2024intrinsic}}} 
    \put(0.74\linewidth,5.8em){\tinytext{\textbf{MultiGP}}}
    \put(0.915\linewidth,5.8em){\tinytext{GT}}
    \end{picture}
  \caption{Additional qualitative results of illumination estimates on the newly captured MultiGP dataset. MultiGP more accurately recovers the illumination structure of the actual environment compared to prior works, including IID \cite{kocsis2024intrinsic}.}
  \label{fig:ourdata_env_addition}
\end{figure}

We additionally compare our method with Kocsis et al. in \cref{tab:results_ourdata_env}. Their Intrinsic Image Diffusion generates texture and reflectance samples, which are then used to optimize illumination. Their lighting is defined by global Spherical Gaussians (SG) \cite{li2020inverse} for incident light and point light sources with SG emission profiles. This parameterization restricts their illumination to low frequencies, as shown in \cref{fig:ourdata_env_addition}, though they still achieve the second-best accuracy in the table. To evaluate their results, we followed their implementation to convert the estimated global SGs and point lights into environment maps by computing their summation along each direction.

\section{FID/NIQE comparison with DPI \cite{lyu2023dpi}}
\cref{tab:FID} shows FID/NIQE comparison with
DPI. DPI scores better on FID because it samples from a learned natural panorama prior—FID measures similarity to a panorama dataset, not to the actual scene. MultiGP outperforms DPI on NIQE (intrinsic image quality, no ref. dist.) on both synthetic and real, indicating DPI’s apparent sharpness is hallucinated detail uncorrelated with the observed scene as shown in \cref{fig:vsDPI}. We did not warm-start from a pretrained natural image/HDR-panorama prior—this would re-inject the hallucinated, scene-uncorrelated detail diagnosed above.

\clearpage
\noindent
\begin{minipage}{\textwidth}
  \centering
  \begin{minipage}[t]{0.50\linewidth}
    \vspace{10pt}
    \centering
    \captionof{table}{FID/NIQE comparison with DPI \cite{lyu2023dpi}. The ``Real'' uses MultiGP dataset. MultiGP outperforms DPI on NIQE.}
    \label{tab:FID}
    \setlength{\tabcolsep}{1pt}
    \setlength{\aboverulesep}{1pt}
    \setlength{\belowrulesep}{1pt}
    \renewcommand{\arraystretch}{0.66}
    \begin{tabular}{l cc cc}
        \toprule
        & \multicolumn{2}{c}{Synthetic} & \multicolumn{2}{c}{Real} \\
        \cmidrule(lr){2-3} \cmidrule(lr){4-5}
        & FID$\downarrow$ & NIQE$\downarrow$ & FID$\downarrow$ & NIQE$\downarrow$ \\
        \midrule
        DPI & \textbf{136.8} & 11.12 & \textbf{367.8} & 12.83 \\
        \textbf{MultiGP} & 246.3 & \textbf{9.08} & 409.7 & \textbf{7.19} \\
        \bottomrule
    \end{tabular}
  \end{minipage}
  \hfill
  \begin{minipage}[t]{0.46\linewidth}
    \vspace{2pt}
    \centering
    \includegraphics[width=\linewidth]{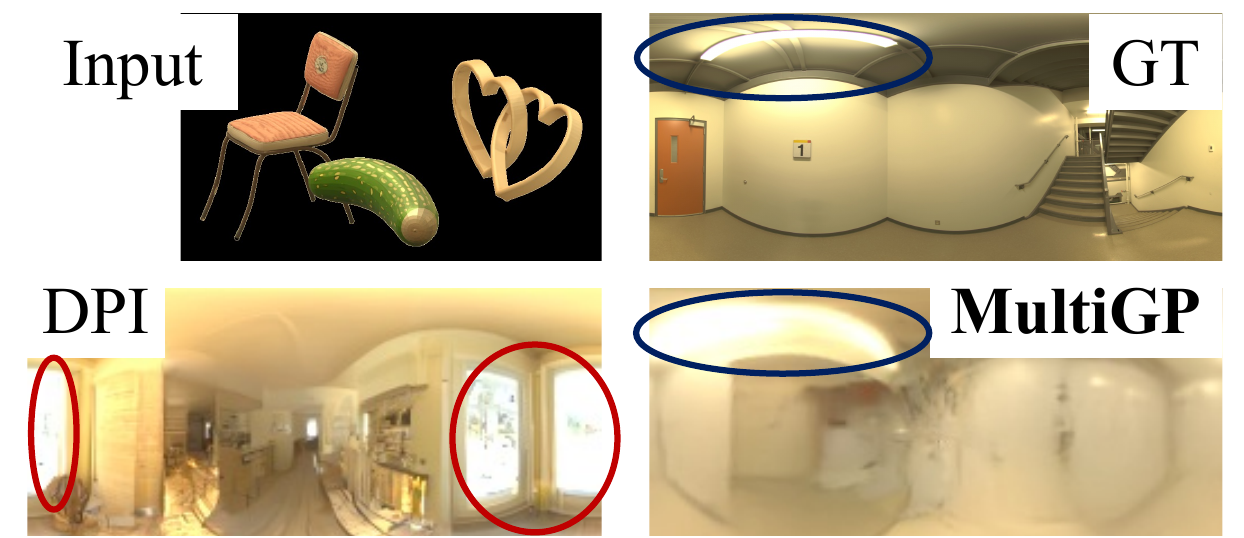}
    \vspace{-16pt}
    \captionof{figure}{Additional qualitative comparison of lighting with DPI \cite{lyu2023dpi}. DPI’s apparent sharpness is hallucinated.}
    \label{fig:vsDPI}
  \end{minipage}
\end{minipage}

\end{document}